\documentclass{article}

\usepackage{PRIMEarxiv}
\AtBeginDocument{\newgeometry{textheight=9in,textwidth=516pt,top=1in,headheight=14pt,headsep=25pt,footskip=30pt}%
  \setlength{\columnsep}{12pt}}

\usepackage[utf8]{inputenc}
\usepackage[T1]{fontenc}
\usepackage{url}
\usepackage{booktabs}
\usepackage{amsfonts}
\usepackage{nicefrac}
\usepackage{microtype}
\usepackage{fancyhdr}
\usepackage{graphicx}
\usepackage{multirow}
\usepackage{array}
\usepackage[colorlinks,urlcolor=blue,linkcolor=blue,citecolor=blue]{hyperref}
\usepackage{amsmath}
\usepackage{amssymb}
\usepackage{mathtools}
\usepackage{amsthm}
\usepackage{xcolor}
\usepackage{algorithm}
\usepackage{algorithmic}
\usepackage{natbib}

\definecolor{wordblue}{HTML}{2F5D9E}
\definecolor{wordorange}{HTML}{CC5C28}
\newcommand{\bv}[1]{\textcolor{wordblue}{\textbf{#1}}}

\renewenvironment{proof}{%
  \par\vspace{1ex}\noindent{\it Proof:}\ %
}{%
  \hfill$\square$\par\vspace{1ex}%
}

\title{Catastrophic Overfitting, Entropy Gap and Participation Ratio: A Noiseless $l^p$ Norm Solution for Fast Adversarial Training}

\author{
  Fares B. Mehouachi\\
  New York University Abu Dhabi \\
  Saadiyat Island \\
  Abu Dhabi, UAE \\
  \texttt{fm2620@nyu.edu} \\
  \And
  Saif E. Jabari \\
  Department of Civil and Urban Engineering \\
  NYU Tandon School of Engineering \\
  New York, USA \\
  \texttt{sej7@nyu.edu} \\
}

\begin{document}
\maketitle
\begingroup
\renewcommand\thefootnote{}
\footnotetext{\hspace{-1.8em}Code: \url{https://github.com/FaresBMehouachi/lpfgsm}}
\endgroup
\begin{abstract}
Adversarial training is a cornerstone of robust deep learning, but fast methods like the Fast Gradient Sign Method (FGSM) often suffer from Catastrophic Overfitting (CO), where models become robust to single-step attacks but fail against multi-step variants. While existing solutions rely on noise injection, regularization, or gradient clipping, we propose a novel solution that purely controls the $l^p$ training norm to mitigate CO.

Our study is motivated by the empirical observation that CO is more prevalent under the $l^{\infty}$ norm than the $l^2$ norm. Leveraging this insight, we develop a framework for generalized $l^p$ attack as a fixed point problem and craft $l^p$-FGSM attacks to understand the transition mechanics from $l^2$ to $l^{\infty}$. This leads to our core insight: CO emerges when highly concentrated gradients, where information localizes in few dimensions, interact with aggressive norm constraints.
By quantifying gradient concentration through Participation Ratio and entropy measures, we develop an adaptive $l^p$-FGSM that automatically tunes the training norm based on gradient information. Extensive experiments demonstrate that this approach achieves competitive robustness without requiring additional regularization or noise injection, providing a novel and theoretically-principled pathway to mitigate the CO problem.
\end{abstract}
\vskip 0.35in
\paragraph{Impact Statement}
As AI models expand, traditional training becomes increasingly expensive, with adversarial training further compounding this cost. Reducing these expenses is crucial for improving access to robust AI models, especially in safety-critical domains like mobility or autonomous driving. While fast adversarial training offers efficiency, it suffers from Catastrophic Overfitting (CO), leaving models vulnerable to sophisticated attacks. Our work introduces a novel mathematical connection between CO and previously unrelated concepts from quantum mechanics (Participation Ratio) and information theory (entropy gap). By quantifying gradient concentration through these metrics, we demonstrate how they track the onset of CO and enable adaptive norm selection. This unexpected bridge between disparate fields yields a computationally efficient solution without requiring noise injection or extra regularization. Beyond mitigating CO, these connections open new theoretical avenues for understanding adversarial robustness. The practical implementation is straightforward to adopt, providing immediate applications for enhancing model security across domains where adversarial robustness is essential.


\twocolumn
\section{Introduction}
\label{sec:intro}
Deep neural networks (DNNs) have become essential in fields like computer vision, natural language processing, and speech recognition \cite{hinton2012deep,lecun2015deep, vaswani2017attention}. Despite their impressive generalization abilities, DNNs are highly vulnerable to adversarial perturbations (subtle input modifications that cause misclassifications) \cite{szegedy2013intriguing, goodfellow2014explaining, 10816724, 10443697}. This vulnerability poses significant risks in critical applications such as autonomous vehicles \cite{koopman2016challenge, koopman2017autonomous, mehouachi2021detection, mehouachi2021detectionb,  wang2023vulnerability,yang2025urban, mehouachiflowmixer, mehouachi2025flowmixer}, healthcare \cite{finlayson2019adversarial}, and finance \cite{fursov2021adversarial, goldblum2020adversarial}.

The discovery of these vulnerabilities has sparked extensive research into enhancing DNN robustness \cite{madry2017towards, fawzi2016robustness, zhang2019theoretically, 10488755, 10765144, 10485642}. Among various defense strategies, adversarial training, which incorporates adversarially perturbed examples during training, has emerged as one of the most effective approaches \cite{goodfellow2014explaining, madry2017towards, 10755032, 10754649}. However, traditional adversarial training using multiple optimization steps is computationally demanding, particularly for large models and high-dimensional data \cite{madry2017towards, 10415302, 10669090}.

Fast single-step adversarial training methods were developed to address this computational challenge. While initially considered less effective, these methods gained renewed attention following the work of Wong et al.~\cite{wong2020fast}, which also revealed a critical phenomenon: Catastrophic Overfitting (CO). During CO, models maintain robustness against single-step attacks but unexpectedly become vulnerable to multi-step adversaries. To counter this, Wong et al.~\cite{wong2020fast} introduced RS-FGSM, adding random perturbations before FGSM, but its effectiveness reduces with larger perturbation radii \cite{andriushchenko2020understanding}. Andriushchenko and Flammarion~\cite{andriushchenko2020understanding} proposed GradAlign, a regularization technique enhancing local linearity at higher computational costs. Methods like ZeroGrad and MultiGrad \cite{golgooni2021zerograd} focused on neutralizing low-normed gradient directions, while De Jorge et al.~\cite{de2022make} improved RS-FGSM by avoiding clipping and using amplified noise. Recent advances explored bi-level optimization perspectives \cite{zhang2022revisiting,wang2024preventing} and prior-guided initialization \cite{jia2022prior}. More recent efforts regularize abnormal adversarial examples~\cite{lin2023eliminating}, analyze layer-wise pseudo-robust shortcuts~\cite{lin2024layer}, and enforce local linearity efficiently~\cite{abadrocamora2024efficient}. Despite these various approaches, which often involve trade-offs such as increased computational overhead or noisier training data, the fundamental mechanism of CO remains elusive.

\begin{figure}[!ht]
  \centering
  \includegraphics[width=0.96\linewidth]{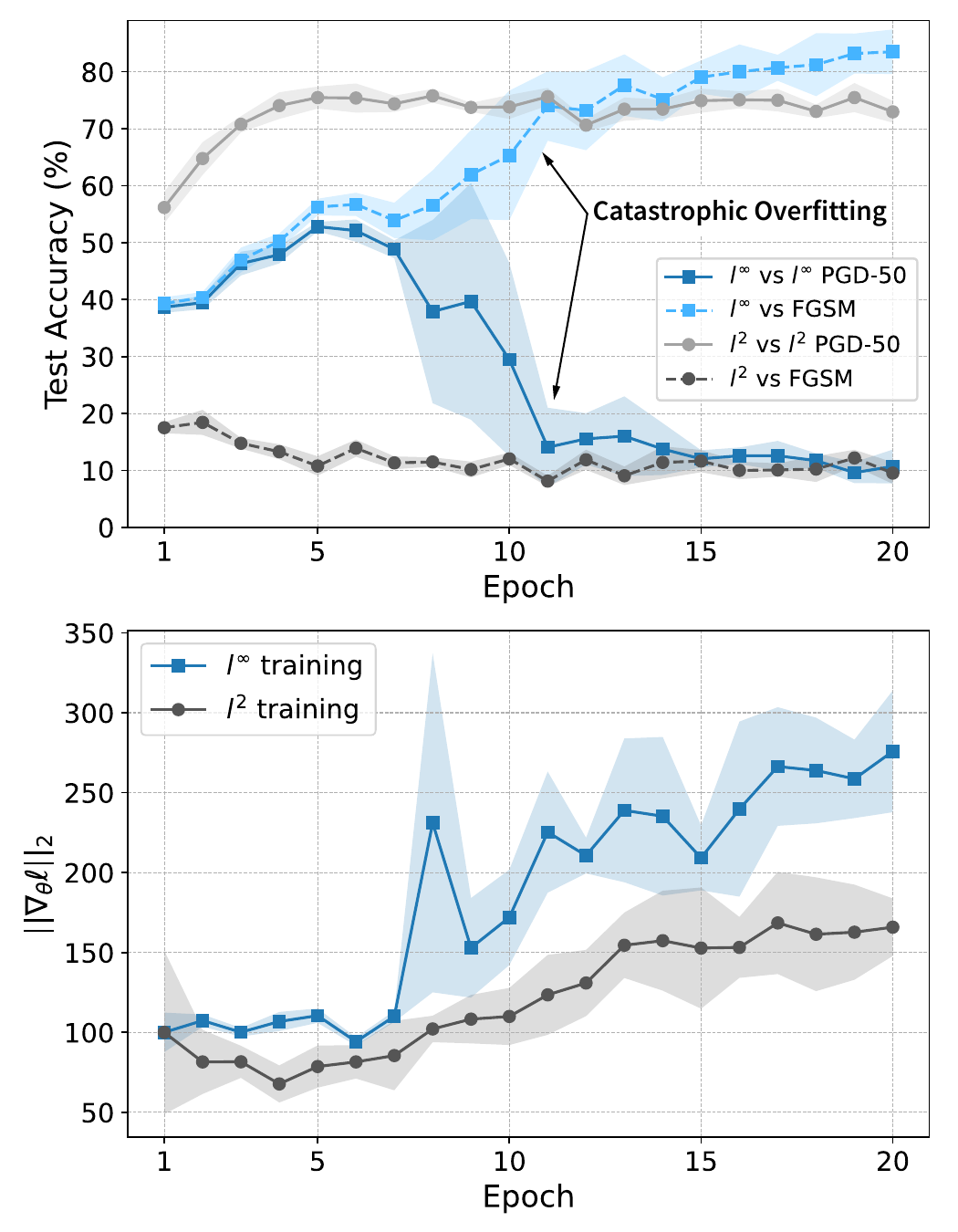}
  \vskip -0.1in
\caption{CO phenomena on CIFAR-10~\cite{krizhevsky2009learning} using WideResNet-28-10~\cite{zagoruyko2016wide}: \textbf{Upper:} $l^{\infty}$ training ($\epsilon = 8/255$) shows accuracy collapse against PGD-50 ($\epsilon = 8/255$)~\cite{madry2017towards} attacks, while $l^2$ ($\epsilon = 32/255$, both training and attack) remains stable. \textbf{Lower:} CO onset in $l^{\infty}$ training correlates with gradient norm increase, absent in $l^2$ training (norms normalized at epoch 1).}
  \label{fig:fig1}
  \vskip -0.10in
\end{figure}

Our work begins with an intriguing observation: CO is predominantly associated with training under the $l^\infty$-norm, while $l^2$-defense remains resistant, albeit with limited robustness to $l^\infty$ attacks (Figure~\ref{fig:fig1}). Moving beyond traditional linear approximations, we reformulate adversarial attack generation as a fixed-point problem and derive the $l^p$-FGSM attack formulation as a single-step optimization.

\begin{figure*}[!ht]
\centering
\includegraphics[width=1.0\linewidth]{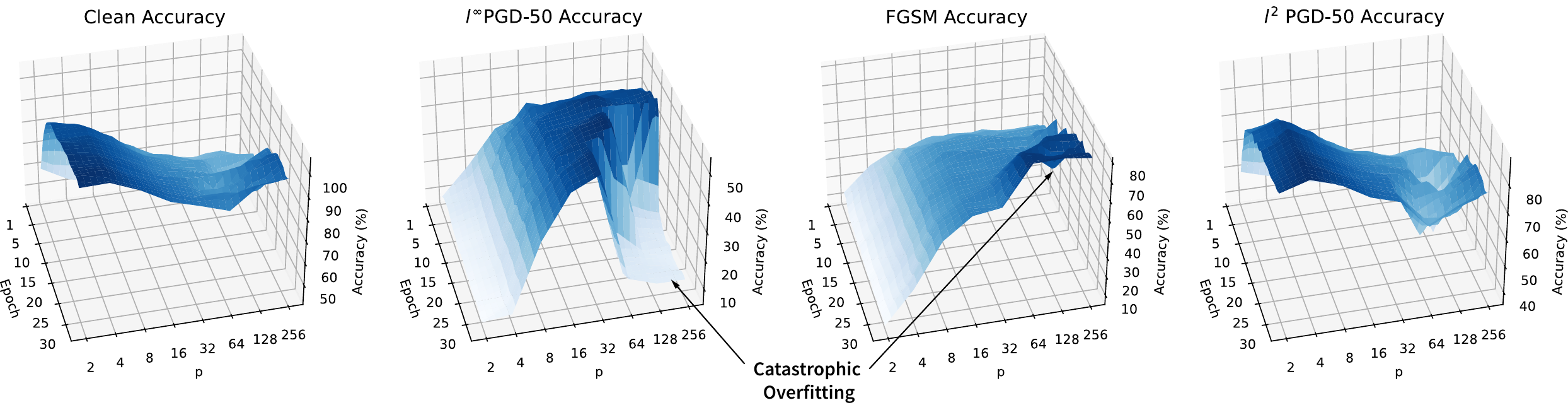}
\caption{Impact of $l^p$ norm choice on training dynamics and robustness for CIFAR-10~\cite{krizhevsky2009learning} with WideResNet-28-10~\cite{zagoruyko2016wide}. The choice of $p$ reveals a key trade-off: higher values ($p \geq 32$) initially show better robustness but become vulnerable to Catastrophic Overfitting (CO)~\cite{kim2021understanding,he2023investigating}, evident in the $l^\infty$ PGD-50~\cite{madry2017towards} plot (second left). Lower $p$ values prevent CO but with reduced adversarial robustness. Notably, $l^2$ PGD-50 accuracy (rightmost) remains stable across different $p$ values, suggesting $l^2$ robustness is less sensitive to norm choice. Results shown for $\epsilon = 8/255$ over 30 epochs. $l^p$-FGSM used for these plots is detailed in Algorithm~\ref{alg:lp-fgsm}.}
\label{fig:fig1x}
\vskip -0.20in
\end{figure*}

Initial exploration of $l^p$-FGSM (Figure~\ref{fig:fig1x}) reveals that higher $p$ values ($p \geq 32$) delay CO but remain susceptible, while lower values prevent CO at the cost of reduced robustness. To resolve this trade-off, we identify gradient concentration as CO's key mechanism, quantified through the Participation Ratio ($\mathtt{PR}$)~\cite{anderson1958absence,feynman2011feynman}, a measure from quantum mechanics of how many components meaningfully contribute to a vector's structure. We adapt $\mathtt{PR}$ to adversarial training by measuring gradient concentration through $\mathtt{PR_1}$, which naturally connects to the angular separation between $l^2$ and $l^\infty$ bounded perturbations. Our adaptive approach selects $p$ based on $\mathtt{PR}$: lower values for concentrated gradients and higher values otherwise, preserving some alignment with the natural $l^2$ geometry.

Further investigation reveals fundamental relationships between $\mathtt{PR_1}$, entropy gap, and norm selection, leading to a principled norm adaptation. Without further noise injection or regularization~\cite{wong2020fast,andriushchenko2020understanding}, our method achieves competitive performance on standard benchmarks solely by adjusting adaptively the value of the adversarial training $p$ norm.

\section{Preliminaries}
\label{sec:preliminaries}

In this paper, we study adversarial robustness in the context of deep learning. We consider, generally, a classification function $c(x; \theta): x \mapsto \mathbb{R}^{C}$, which transforms input features $x$ into output logits associated with classes in set $C$. The probability $\pi_{i}(x; \theta)$ of predicting label $i$ for input $x$ is defined through a softmax function: $\exp \left(c_{i}(x; \theta)\right) / \sum_{j} \exp \left(c_{j}(x; \theta)\right)$, where $c_{i}(x; \theta)$ is the $i$-th element of the output logits and $\theta$ denotes the model parameters \cite{goodfellow2016deep}. Adversarial robustness, in terms of the function $c$, is characterized as follows: the function $c$ is deemed robust to adversarial perturbations of magnitude $\epsilon$ at input $x$ if, and only if the class with the maximum probability for input $x$ retains the highest probability for the input $x+\delta$, where $\delta$ is any adversarial perturbation confined within the $l^p$ ball of radius $\epsilon$~\cite{szegedy2013intriguing, goodfellow2014explaining}. This concept can be succinctly formulated as follows:
\begin{equation}
   \underset{i \in C}{\operatorname{argmax}}\, \pi_{i}(x+\delta; \theta)= \underset{i \in C}{\operatorname{argmax}}\, \pi_{i}(x; \theta), \: \forall \delta \in B_{p}(\epsilon).
\end{equation}
This study focuses on instances where the norm extends beyond $l^{2}$ or $l^{\infty}$ and could be any $l^p$ with $p\geq 2$. For the sake of simplicity, $B(\epsilon)$ is employed to represent $B_{p}(\epsilon)$. Given a dataset with a distribution $\mathcal{D}$, the prevalent approach for training a classifier $c$ is through Empirical Risk Minimization (ERM) \cite{vapnik1999nature}:
\begin{equation}
\min_{\theta} \mathbb{E}_{(x, y) \sim \mathcal{D}}[\ell(x; y, \theta)].
\end{equation}
where
$\ell$ is a loss function, often the standard cross-entropy  $\ell\left(x;y,\theta\right)=-y^{T}\log\left(\pi\left(x;\theta\right)\right)$, and $y$ is a one-hot encoded vector that describes the class label. Despite ERM's proven effectiveness in training neural networks to attain satisfactory performance on unseen data, it falls short in the face of adversarial attacks \cite{szegedy2013intriguing, goodfellow2014explaining}. The shift in the data distribution created by the attacks causes the test accuracy to drop substantially. To address this shortcoming and enhance the network's robustness, adversarial training \cite{goodfellow2014explaining, madry2017towards} is typically used. This approach uses crafted adversarial attacks for the training to simulate potential distributional shifts. Such a strategy steers the model to learn features that remain robust to minor input perturbations. The injection of the adversarial training inside the loss function could be expressed as follows:

\begin{equation}
\min_{\theta}\, \mathbb{E}_{(x, y) \sim \mathcal{D}}\left[\max _{\delta \in B(\epsilon)} \ell(x+\delta; y, \theta)\right].
\label{eq:adv}
\end{equation}

In the equation, the inner maximization, $\max _{\delta \in B(\epsilon)} \ell(x+\delta; y, \theta)$, is typically carried out through a set number of steps employing a gradient-based optimization technique. Projected Gradient Descent (PGD)~\cite{madry2017towards} is a prevalent method that entails the subsequent update:

\begin{equation}
\delta \leftarrow \operatorname{\Pi}\left(\delta+\mu \nabla_{x} \ell(x+\delta; y, \theta)\right).
\end{equation}

The projection operator $\operatorname{\Pi}$, taking the form of either scaling or truncation for $l^2$ and $l^\infty$, respectively, ensures perturbations remain within the predefined bounds. Employing multiple steps to craft the adversarial perturbation can rapidly escalate computational expenses. A more economical strategy for adversarial training employs a linear approximation of the loss function $\ell(x_{0}+\delta)\approx\ell(x_{0})+\delta^{T}\nabla_{x}\ell$, and adopts gradient sign as a solution to the maximization problem. This strategy, known as the Fast Gradient Sign Method (FGSM) \cite{goodfellow2014explaining}, provides computational efficiency, albeit potentially generating sub-optimal adversarial examples.
\begin{equation}
\delta_{\textit{FGSM}}=\underset{\delta\in B_{\infty}(\epsilon)}{\operatorname{argmax}}\,\left(\ell(x_{0})+\delta^{T}\nabla_{x}\ell\right)=\epsilon\,\text{sign}\left(\nabla_{x}\ell\right).
\end{equation}

The FGSM perturbation, being facile to compute, precisely resolves the linearized maximization problem~(\ref{eq:adv}) under $l^{\infty}$ constraint. However, as observed by~Wong et al.~\cite{wong2020fast}, FGSM suffers from Catastrophic Overfitting, prompting the proposition of random noise addition $\eta\sim\mathcal{U}\left[-\epsilon,\epsilon\right]$ as a remedy for the CO issue.
\begin{equation}
\delta_{\textit{RS-FGSM}}=\Pi_{B_{\infty}(\epsilon)}\left(\eta+\epsilon\,\text{sign}\left(\nabla_{x}\ell\left(x_{0}+\eta\right)\right)\right).
\end{equation}

Our work focuses on characterizing the inner maximization in~(\ref{eq:adv}) beyond first-order approximations using an $l^p$ constraint, yielding a fixed point formulation.

\section{Theoretical Framework}
\label{sec:theory}

In this section, we relax the local linearity assumption (first-order Taylor expansion) commonly employed in FGSM by considering local convexity. Through empirical evidence, we show that local convexity emerges naturally during training, offering deeper insights into the geometry of adversarial perturbations.\footnote{If local convexity does not hold, the framework can default to local linearity.} \footnote{For one-step adversarial training, local linearity and convexity lead to identical outcomes.} This perspective reveals that optimal perturbations reside on the boundaries of permissible constraints, allowing us to formulate the problem using a fixed-point approach. Starting with the $l^2$ case, which highlights connections to GradAlign, we extend this framework to general $l^p$ norms and develop an adaptive norm selection mechanism based on gradient concentration metrics.

\subsection{Local Convexity and Attacks Optimality}
\begin{figure}[!ht]
  \centering
   \includegraphics[width=1.0\linewidth]{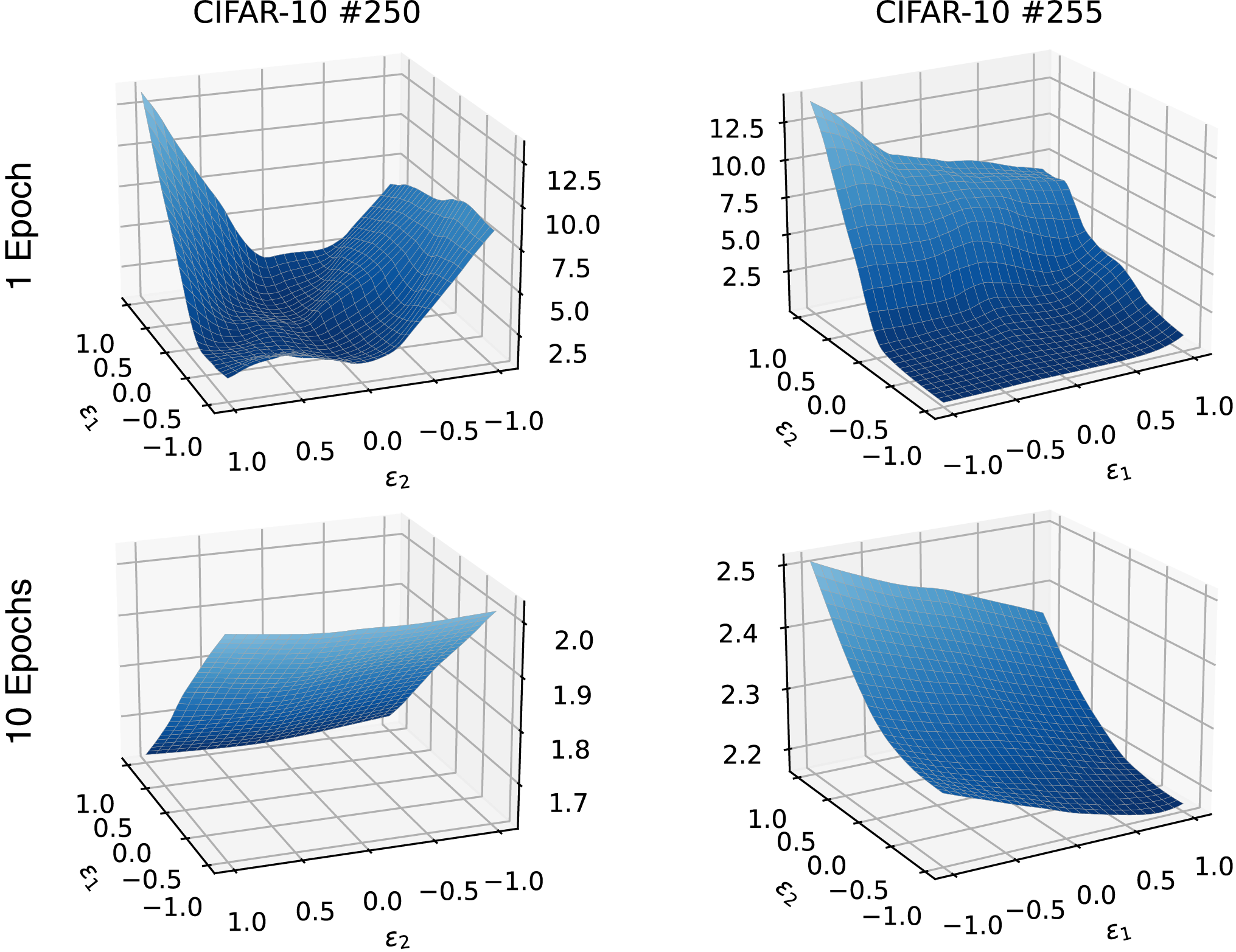}
      \caption{Depiction of training effect on CIFAR-10's loss landscape at different training points. The upper panels display the landscape after one epoch and the lower ones ten epochs with $l^p$-FGSM (Alg.~\ref{alg:lp-fgsm}). Training points are positioned at $(0,0)$, $\varepsilon_1$ and $\varepsilon_2$ are eigenvectors corresponding to the Hessian's ($\nabla^{2}_x \ell$) extreme eigenvalues for each sample. Training induces local convexity.}
   \label{fig:fig2x}
   \vskip -0.3cm
\end{figure}

While fast adversarial training traditionally relies on local linearity assumptions, we examine a local convexity framework that emerges from analyzing the Hessian of the loss function with respect to inputs. When the Hessian $\nabla^2_x \ell$ is positive ($\nabla^2_x\ell\succeq 0$) on the perturbation ball, the loss is convex there and its maximum occurs on the boundary $\partial B_p(\epsilon)$ - a property that enables efficient single-step methods. While the Hessian may not be positive at initialization, we empirically observe (Figure~\ref{fig:fig2x}) that local convexity emerges reasonably within a few epochs of training. This justifies our assumption for the analysis of adversarial perturbations during the main training phase. Importantly, when convexity does not hold (typically in early training), our framework defaults to the local linearity assumption, reducing to standard FGSM behavior. The Hessian decomposition~\cite{mehouachi2021exact, mehouachi2024sifrian, mehouachi2024sifr}, with respect to the output logits, reveals:
\begin{equation}
\nabla_{x}^{2}\ell=\left(\frac{\partial \pi}{\partial x_{0}}\right)\frac{\partial^2\ell}{\partial \pi^2}\left(\frac{\partial \pi}{\partial x_{0}}\right)^{T}+\frac{\partial^{2}\pi}{\partial x_{0}^{2}}\frac{\partial\ell}{\partial \pi}.
\end{equation}
This structure combines a positive Gauss-Newton term (which dominates as training progresses) with a second term that diminishes during training as errors $\frac{\partial\ell}{\partial \pi}$ decrease. While this convergence to positive curvature can be accelerated by controlling $\frac{\partial^{2}\pi}{\partial x_{0}^{2}}$ through architectural choices like SELU~\cite{klambauer2017self} or GELU~\cite{hendrycks2016gaussian} activations, our empirical analysis shows that even standard ReLU networks develop local convexity through training, as visualized in Figure~\ref{fig:fig2x}. This observation provides theoretical justification for boundary-focused search strategies while relaxing the local linearity assumption.
\subsection{$l^2$ Norm-Bounded Adversarial Attacks}
Given the local convexity of $\ell$, the optimal perturbation exists on the boundary. Using a Lagrange multiplier, we reformulate the maximization problem~(\ref{eq:adv}) as unconstrained.

\noindent\textbf{Proposition 1.} \emph{
For a training sample $x_0$ with non-null gradient, the optimal perturbation $\delta^\star$ within $B(\epsilon)$ exists and solves the fixed-point problem $\delta^{\star}=F(\delta^{\star})$, where
\begin{equation}
F(\delta)=\epsilon\frac{\nabla_{x}\ell(x_{0}+\delta)}{\|\nabla_{x}\ell(x_{0}+\delta)\|_2}.
\label{eq:fx2}
\end{equation}
$F$ is Lipschitz around its origin with constant $K=2\epsilon\|\nabla_{x}^{2}\ell\|/\|\nabla_{x}\ell(x_{0})\|_2$:
\begin{equation}
\|F(\delta)-F(0)\| \leq K\|\delta\|,
\end{equation}
and the fixed-point problem converges if $K<1$.}

\noindent\textbf{Proof.} See Supplementary Materials Section I.%

\begin{figure}[ht]
\centering
\includegraphics[width=0.98\linewidth]{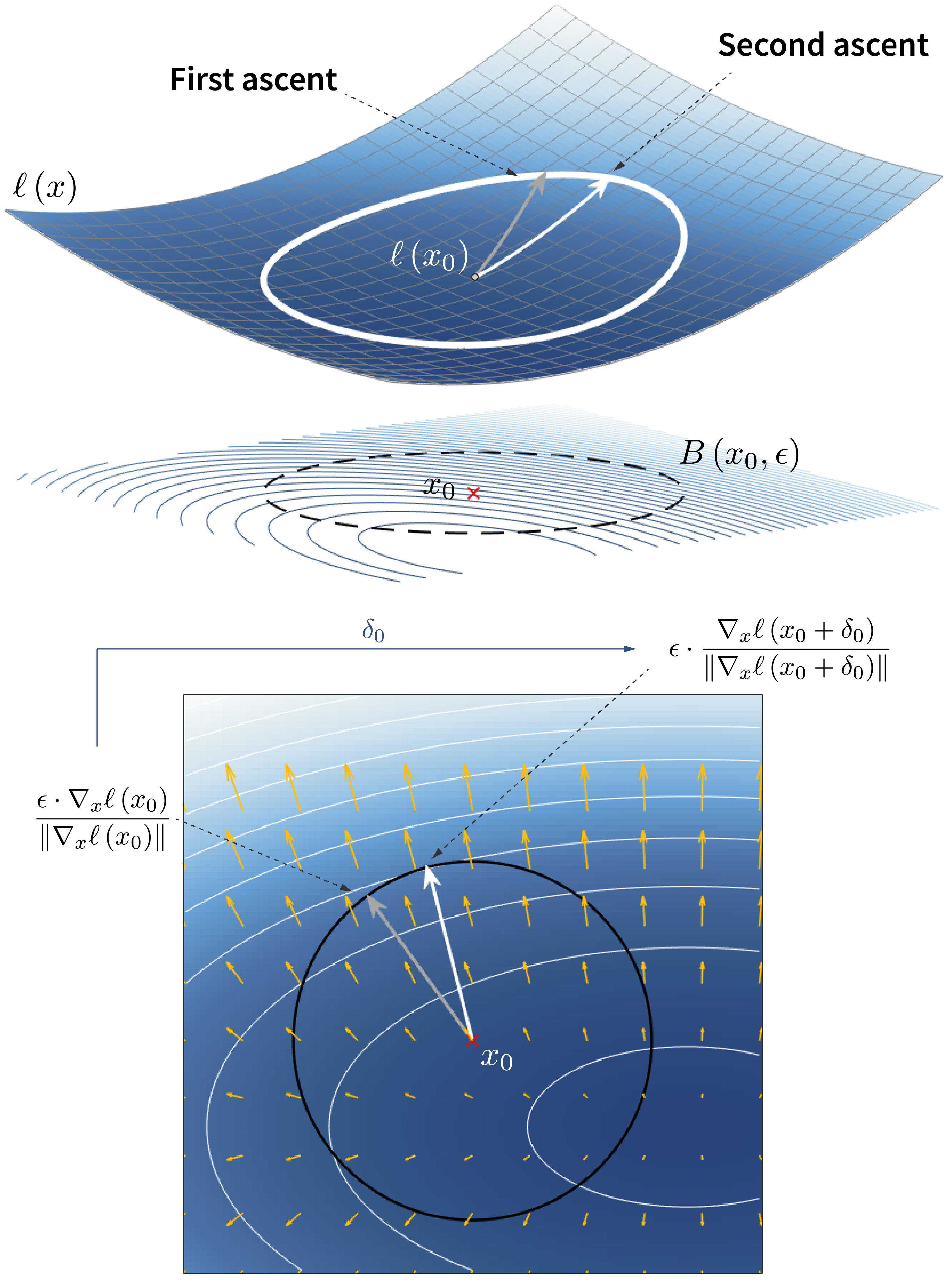}
\caption{Illustration of the initial two ascents of the fixed-point algorithm~(\ref{eq:fx2}) for optimal perturbation identification under the $l^2$ constraint.}
\label{fig:fig3}
\end{figure}

Equation~(\ref{eq:fx2}) defines a fixed-point problem that iteratively approximates the optimal perturbation, as shown in Figure~\ref{fig:fig3}. While CURE~\cite{moosavi2019robustness} minimizes Hessian norm for robustness, Srinivas et al.~\cite{srinivas2022efficient} introduced gradient norm division for scale-invariant curvature, which we identify as our Lipschitz constant $K$. Reducing $K$ accelerates inner maximization convergence~(\ref{eq:adv}). The fixed-point convergence also provides insight into GradAlign~\cite{andriushchenko2020understanding}.

\noindent\textbf{Corollary (GradAlign).} \emph{
When $\nabla_x \ell(x_0)$ aligns with $\nabla_{x}\ell(x_{0}+\epsilon\nabla_{x_0}\ell/\|\nabla_{x_0}\ell\|)$, the fixed-point converges instantly}\footnote{In this ideal case, the bounded gradient is the fixed point.}:
\begin{equation}
\frac{\nabla_{x}\ell(x_{0}+\epsilon\nabla_{x_{0}}\ell/\|\nabla_{x_{0}}\ell\|)}{\|\nabla_{x}\ell(x_{0}+\epsilon\nabla_{x_{0}}\ell/\|\nabla_{x_{0}}\ell\|)\|}=\frac{\nabla_{x_{0}}\ell}{\|\nabla_{x_{0}}\ell\|}.
\end{equation}

GradAlign~\cite{andriushchenko2020understanding} regularizes gradient alignment, effectively improving the initial point of our fixed-point algorithm.

\subsection{$l^p$ Norm-Bounded Adversarial Attacks}
\label{sec:novelty}
The $l^p$ norm serves as a smooth proxy to $l^\infty$ as $p$ increases. Following our $l^2$ analysis with Lagrange multipliers, we characterize $l^p$ optimal attacks as a fixed-point problem:

\noindent\textbf{Proposition 2.} \emph{
For a training sample $x_0$ with non-null gradient under a $B_p(\epsilon)$ constraint, the optimal perturbation $\delta^\star$ exists and solves the fixed-point equation $\delta^{\star}=F_p(\delta^{\star})$, where:
\begin{equation}
F_{p}(\delta)=\epsilon\,\text{sign}(\nabla_{x}\ell(x_{0}+\delta))\left|\frac{\nabla_{x}\ell(x_{0}+\delta)}{\|\nabla_{x}\ell(x_{0}+\delta)\|_{q}}\right|^{q-1},
\label{eq:fxp}
\end{equation}
with $l^q$ being the dual norm of $l^p$: $\frac{1}{p}+\frac{1}{q}=1$. The absolute value and multiplication operations are element-wise.}

\noindent\textbf{Proof.} See Supplementary Materials Section II.%

\begin{figure}[ht]
\centering
\includegraphics[width=0.9\linewidth]{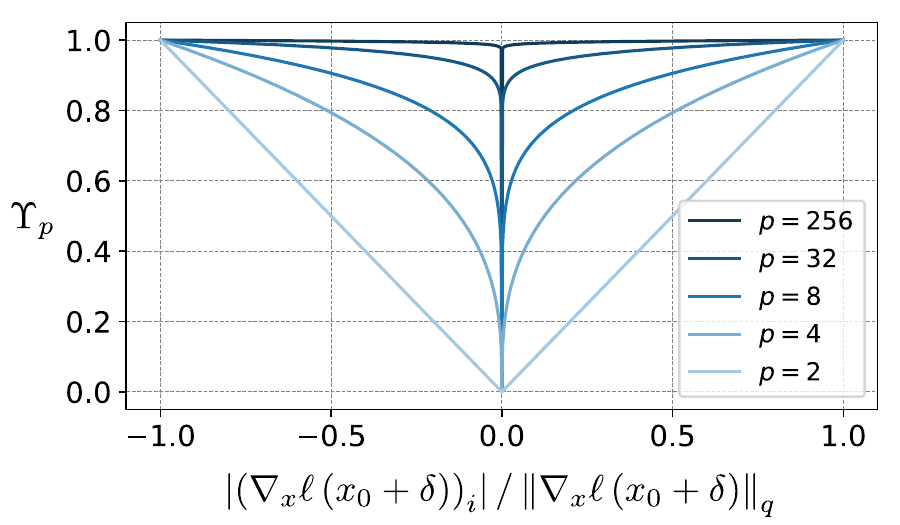}
\caption{Variation of the $l^p$ transition function $\Upsilon_p$ for different values of $p$. The high-pass filtering effect mirrors the thresholding behavior in ZeroGrad~\cite{golgooni2021zerograd}.}
\label{fig:fig4}
\end{figure}

\vspace{0.1cm}
\noindent\textbf{The $l^p$ attack spectrum from $l^2$ to $l^\infty$:}
The formula (\ref{eq:fxp}) for $l^p$ optimal attacks is valid for any $p\geq 2$, for $p=q=2$, we get the same formula (\ref{eq:fx2}), while for $p\rightarrow +\infty$ we get $q=1$ and we find the same formula as FGSM~\cite{goodfellow2014explaining}. Furthermore it is straightforward to verify that $\|F_{p}(\delta)\|_{p}=\epsilon,$ since $p(q-1)=q.$
For any $q>1$, the single-step $l^p$ perturbation remains continuous with respect to $\nabla_x \ell$, even as the gradient approaches zero. The transition between $l^\infty$ and $l^2$ is governed by:
\begin{equation}
\Upsilon_{p}(\delta)=\left|\frac{\nabla_{x}\ell(x_{0}+\delta)}{\|\nabla_{x}\ell(x_{0}+\delta)\|_{q}}\right|^{q-1}.
\end{equation}
which acts as a high-pass filter, approaching unity everywhere except near zero (Figure~\ref{fig:fig4}).

\noindent\textbf{Lipschitzness of $F_p$:} For $p > 2$, global Lipschitz continuity fails due to the discontinuous sign function and concave power term $q - 1$ at null gradients. However, local Lipschitzness suffices via Banach contraction when gradients are bounded away from zero:
\begin{equation}
\exists\, m > 0: \forall i, \forall \delta \in \partial B_p(\epsilon), |\nabla_{x} \ell (x_0 + \delta)_i| > m.
\label{eq:stringent}
\end{equation}
Under this and Proposition 1 conditions, $F_p$ is locally Lipschitz around the origin: $\exists K(p,m)\geq0$ such that:
\begin{equation}
\|F_{p}(\delta)-F_{p}(0)\| \leq K(p,m)\epsilon\frac{\|\nabla_{x}^{2}\ell\|}{\|\nabla_{x}\ell(x_{0})\|_{q}}\|\delta\|.
\end{equation}
The explicit form of $K(p,m)$ is in  Supplementary Materials Section III. %
\vspace{0.1cm}
We leverage this insight to ensure both numerical stability and Lipschitzness by adding a small constant $\varepsilon$ to the absolute value of gradients (Algorithm~\ref{alg:lp-fgsm}).
\subsection{Gradient-Aware Norm Selection}
\label{sec:grad_aware}

While $l^p$-FGSM with fixed $p$ values can balance robustness and stability, this approach faces inherent limitations. As $p$ increases, adversarial robustness improves until CO occurs abruptly, forcing us to settle for lower $p$ values that yield suboptimal robustness (Figure~\ref{fig:fig1x}). This sensitivity to $p$ motivates a deeper examination of the relationship between norm selection and gradient behavior.

In a high-dimensional space $\mathbb{R}^d$, the perturbation amplitude depends essentially on the input dimension $d$:
\begin{equation}
\|\delta_{2}\|_{2}=\epsilon,\;\|\delta_{\infty}\|_{2}\stackrel{a.s.}{=}\epsilon\,d^{\frac{1}{2}},\;\max\|\delta_{p}\|_{2}=\epsilon\,d^{(\frac{1}{2}-\frac{1}{p})}.
\end{equation}

These maximal norms\footnote{For $p>2$, maximum occurs when all components have the same amplitude.}, which also appear in adversarial PAC-Bayes bounds~\cite{xiao2023pac}, reveal that $l^{\infty}$-bounded perturbations can yield vectors dramatically far from the original sample as dimension increases. This effect is particularly significant given that even modest image datasets operate in high dimensions: CIFAR-10 yields $d = 32 \times 32 \times 3 = 3,072$, while ImageNet has $d \sim 1.5 \times 10^5$.

Our key insight is that decreasing the norm $p$ effectively reduces the dimensionality of the perturbation space from $d$ to an effective dimension $d_e$. This relationship can be intuitively captured by the approximation: $d^{(\frac{1}{2}-\frac{1}{p})} \sim d_{e}^{\,\frac{1}{2}}$. This suggests a natural path forward: if we can measure the intrinsic effective dimension of a gradient, we can potentially determine an appropriate $p$ value that balances robustness and stability.

A natural measure of effective dimensionality exists in quantum mechanics, where the Participation Ratio ($\mathtt{PR}$)~\cite{anderson1958absence,feynman2011feynman} quantifies electron localization:
\begin{equation}
\mathtt{PR}(x)=\frac{(\sum_{i}|x_{i}|^{2})^{2}}{\sum_{i}|x_{i}|^{4}}=\frac{\left\langle x^{2},\mathbf{1}\right\rangle ^{2}}{\left\langle x^{2},x^{2}\right\rangle }=\left(\frac{\|x\|_{2}}{\|x\|_{4}}\right)^{4}.
\end{equation}

The $\mathtt{PR}$ measures how many components meaningfully contribute to a vector's structure, providing an effective dimensionality bounded between 1 and $d$. Adapting this concept to adversarial training, we substitute the ones vector with the sign vector of gradient, yielding an analogous measure of dimensionality:
\begin{equation}
\mathtt{PR_1}\left(\nabla_{x}\ell\right)=\frac{\left\langle \nabla_{x}\ell,\text{sign}\left(\nabla_{x}\ell\right)\right\rangle ^{2}}{\left\langle \nabla_{x}\ell,\nabla_{x}\ell\right\rangle }=\left(\frac{\|\nabla_{x}\ell\|_{1}}{\|\nabla_{x}\ell\|_{2}}\right)^{2}.
\end{equation}

This effective dimension varies between $1$ and $d$ for non-null vectors and naturally connects to the geometric relationship between $\delta_2$ and $\delta_\infty$ attacks through their angular separation:
\begin{equation}
\cos(\theta_{2,\infty})=\frac{\|\nabla_{x}\ell\|_{1}}{\|\nabla_{x}\ell\|_{2}d^{\frac{1}{2}}}=\sqrt{\frac{\mathtt{PR_1}}{d}}.
\label{eq:cos2inf}
\end{equation}

\begin{figure}[!h]
\centering
\includegraphics[width=1.0\linewidth]{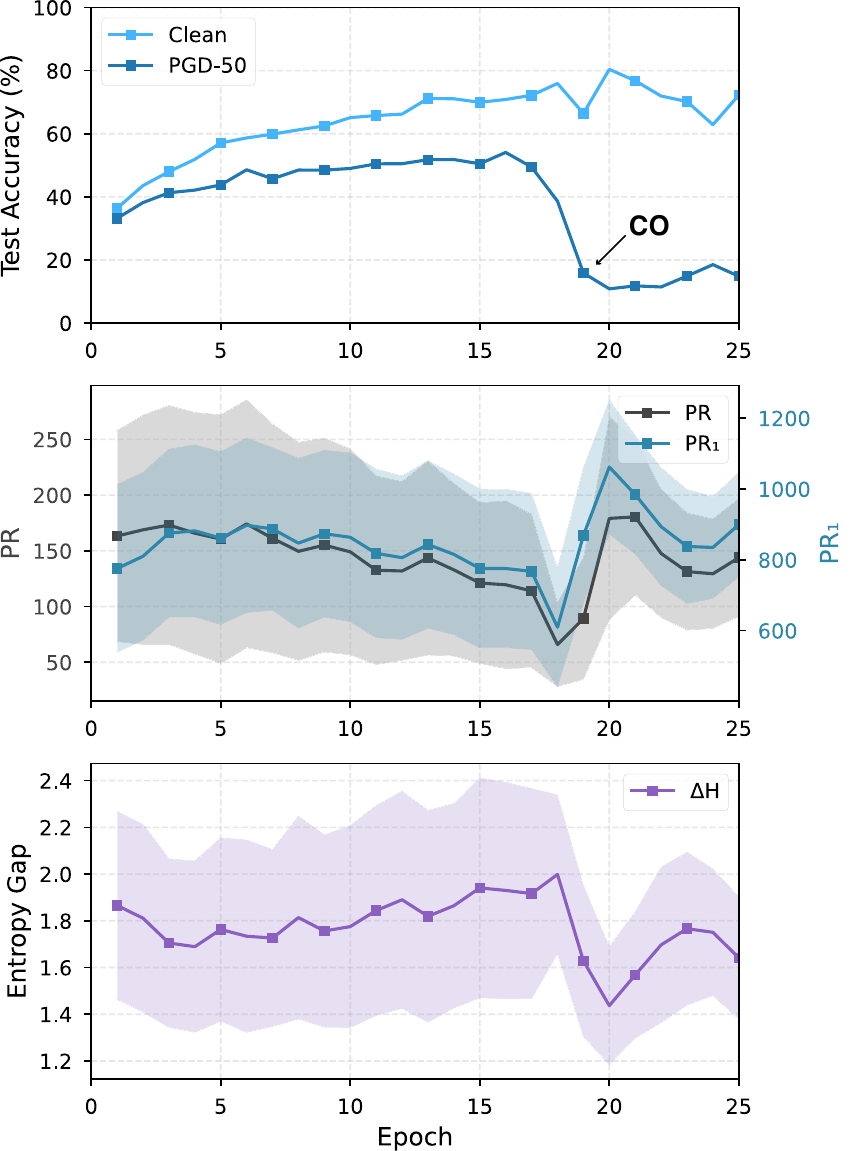}
\caption{Evolution of Participation Ratios ($\mathtt{PR}$, $\mathtt{PR_1}$) and entropy gap during training. Sharp declines in these metrics align with the onset of Catastrophic Overfitting (CO), highlighting the link between gradient concentration and adversarial vulnerability. Shaded areas correspond to $\pm$ standard deviations within the test dataset. Experiment conducted with CIFAR-10, WRN-28-10, ADAM with 0.001 learning rate and a batch size of 128. }
\label{fig:fig6x}
\vskip -0.1in
\end{figure}

Our analysis suggests a key hypothesis: CO emerges when highly concentrated gradients (indicated by low participation ratios $\mathtt{PR}$, $\mathtt{PR}_1$) interact with aggressive $l^\infty$ bounds. This interaction manifests as increasing gap/angle between the gradient and its sign vector, creating vulnerabilities that multi-step attacks can exploit. Figure~\ref{fig:fig6x} provides empirical validation - both participation ratios drop sharply at CO onset, with corresponding increases in angular separation, confirming gradient concentration's role in triggering catastrophic behavior.

Noise injection has been used to mitigate CO~\cite{wong2020fast,de2022make}. The following lemma gives a condition under which additive gradient noise increases $\mathtt{PR_1}$, and hence $\cos(\theta_{2,\infty})$ in \eqref{eq:cos2inf}.

\begin{samepage}
\noindent\textbf{Lemma 1} (Noise-Induced Alignment). \emph{Let $g\in\mathbb{R}^d\setminus\{0\}$, $M>0$, $\eta\sim\mathcal{U}[-M,M]^d$, and
\[
\kappa_M=\frac{1}{dM^2}\sum_{i=1}^{d}(M-|g_i|)_+^2,\qquad \nu_M=\frac{M\|g\|_1}{\|g\|_2^2}.
\]
Then $\mathbb{E}\|g+\eta\|_1=\|g\|_1+\tfrac{dM}{2}\kappa_M$ and $\mathbb{E}\|g+\eta\|_2^2=\|g\|_2^2+\tfrac{dM^2}{3}$. If $\kappa_M>\nu_M/3$, then
\begin{equation}
\mathbb{P}\big[\mathtt{PR_1}(g+\eta)\ge\mathtt{PR_1}(g)\big]\ge1-\exp\!\left(-\frac{d\,(\kappa_M-\nu_M/3)^2}{8\,(\sin\theta_{2,\infty}+\nu_M)^2}\right),
\label{eq:lemma1}
\end{equation}
where $\theta_{2,\infty}$ is the angle \eqref{eq:cos2inf} of $g$.}

\noindent\textbf{Proof.} See Supplementary Materials Section IV. \hfill$\square$
\end{samepage}

Here $\kappa_M\in[0,1]$ is a weighted fraction of the coordinates that the noise overpowers ($|g_i|<M$), and $\nu_M$ compares $M$ with the $l^1$-weighted mean magnitude $\|g\|_2^2/\|g\|_1$. Once $\kappa_M>\nu_M/3$, that is, once the noise overpowers enough small coordinates relative to its own level, it increases $\mathtt{PR_1}$ with a probability that approaches one as $d$ grows, by \eqref{eq:lemma1}, provided $\kappa_M-\nu_M/3$ stays bounded away from zero. Concentrated gradients, which accompany CO (Figure~\ref{fig:fig6x}), have many such coordinates. Noise weaker than every coordinate instead lowers the expected $\mathtt{PR_1}$ to leading order in $M$ (Supplementary Materials Section IV). Lemma~2 shows that lowering $p$ improves the alignment without noise.

\noindent\textbf{Lemma 2} (Monotonicity of Angular Separation). \emph{For any non-null gradient $\nabla_x \ell$ and $p \geq 2$ with dual exponent $q$ where $1/p + 1/q = 1$, let
\begin{equation}
\cos(\theta_{2,p})=\frac{\|\nabla_{x}\ell\|_{q}^{q}}{\|\nabla_{x}\ell\|_{2}\|\nabla_{x}\ell\|_{2(q-1)}^{q-1}}
\label{eq:cos2p}
\end{equation}
be the cosine between $l^2$ and $l^p$ perturbations. Then:
\begin{equation}
\cos(\theta_{2,\infty}) \leq \cos(\theta_{2,p})
\end{equation}
Equivalently, $\theta_{2,p} \leq \theta_{2,\infty}$; more generally, $\theta_{2,p}$ increases monotonically with $p$.}

\noindent\textbf{Proof.} See Supplementary Materials Section V. \hfill$\square$

\begin{figure}[ht]
\centering
\includegraphics[width=0.9\linewidth]{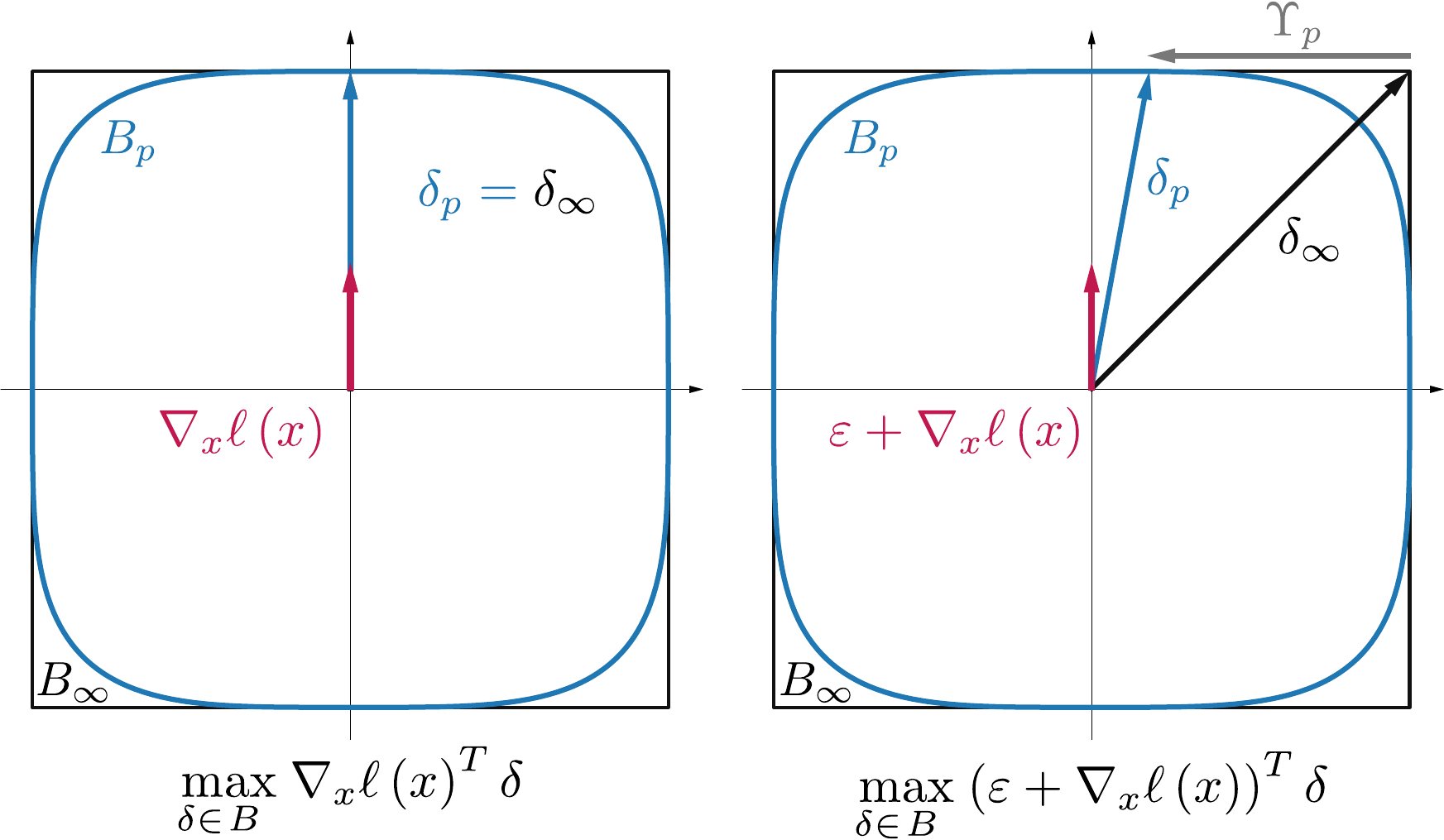}
\caption{Effect of the $l^p$ norm on attack geometry and sensitivity to gradient noise. \textbf{Left:} An ideal scenario, where the angles between $\delta_2$, $\delta_\infty$, and any $\delta_p$ are zero. \textbf{Right:} Under small gradient noise (common in ML), $l^\infty$ shows high sensitivity with large angular separation, whereas $l^p$ yields more stable attacks with better gradient alignment (higher cosine similarity).}
\label{fig:fig2}
\vskip -0.1 cm
\end{figure}

This monotonicity property suggests a natural defense strategy: adaptively choosing the norm based on gradient structure. When gradients concentrate, shifting from higher norms to safer, lower $p$ values aligns better with the natural $l^2$ geometry of the loss landscape, offering a balance between robustness and stability. However, directly determining a suitable $p$ value from (\ref{eq:cos2p}) is challenging in practice.

Considering that $q \in [1,2]$ and aiming for moderate increase in $q$, a first-order Taylor expansion provides a more computationally efficient approach\footnote{Details are in Supplementary Materials Section VI}:%

\begin{equation}
\cos(\theta_{2,p})=\sqrt{\frac{\mathtt{PR}_1}{d}}\left(1+(q-1)\Delta H\right)+\mathcal{O}((q-1)^{2})
\end{equation}

where $\Delta H=H_{m}-H$ is the Entropy Gap, $H$ is the Shannon entropy of the normalized gradient components:
\begin{equation}
H = -\sum_{i=1}^d \rho_i\log(\rho_i), \quad \rho_i = \frac{|\nabla_x\ell_i|}{\|\nabla_x\ell\|_1}
\end{equation}
and $H_m$ is the logarithmic mean entropy:
\begin{equation}
H_m = -\log\prod_{i=1}^d(\rho_i)^{\frac{1}{d}}
\end{equation}

The entropy gap $\Delta H = H_m - H$ is non-negative by Jensen's inequality. If we insert a barrier threshold $\tau$, below which the cosine should not drop $\cos(\theta_{2,\infty})\leq\tau\leq\cos(\theta_{2,p})$, then we can derive the following threshold for the $p(q)$ norm value:
\begin{equation}
q^{*}\geq1+\frac{(\tau\sqrt{\frac{d}{\mathtt{PR_1}}}-1)}{\Delta H},\;\tau\in[0,1]
\label{eq:qstar}
\end{equation}

This formula captures the interplay between gradient geometry, information measures, and norm selection: at the onset of CO, gradients concentrate (low $\mathtt{PR_1}$), the entropy gap $\Delta H$ decreases (Figure~\ref{fig:fig6x}), driving $q$ toward higher values (lower $p$) to maintain alignment. Conversely, well-distributed gradients yield $q$ close to 1, allowing higher $p$ values for enhanced robustness. The threshold $\tau$ serves as a single, interpretable hyperparameter, representing a critical separation angle to balance this trade-off. For practical use, $\tau$ can be also defined as:
\begin{equation}
\tau \equiv \cos((1 - \beta)\theta_{2,\infty}).
\label{eq:hyper}
\end{equation}
This $\beta$ parameterization provides seamless continuity
across the $l^p$ spectrum: $\beta = 0$ preserves $l^\infty$ geometry while
$\beta = 1$ targets $l^2$ alignment, within the first-order approximation behind \eqref{eq:qstar}. In practice, we recommend $\beta \in [0.01, 0.1]$, and it naturally represents the relative reduction of the separation angle.
\subsection{$l^p$-FGSM Algorithm}

Our $l^p$-FGSM algorithm maximizes a locally convex/linear loss under $l^p$-norm bound through one fixed-point iteration ($\delta^{(1)} = F_p(\delta^{(0)})$) with zero initialization. The key innovation is the adaptive norm selection mechanism that automatically adjusts $p$ based on gradient concentration metrics during training, for each sample in the batch individually.

\begin{algorithm}[!th]
   \caption{$l^p$-FGSM}
   \label{alg:lp-fgsm}
\begin{algorithmic}
\STATE {\bfseries Input:} Model $\theta$, data $x$, labels $y$, loss $\ell$, optimizer, attack amplitude $\epsilon$, norm $p(q)$.
   \REPEAT
   \STATE Sample minibatch $(x_0, y_0)$
    \STATE Compute gradient $g_{x} \leftarrow \nabla_{x_0}\ell(x_0 ,y_0)$;
    \STATE Ensure Stability / Lipschitzness $\bar{g}_{x}\leftarrow\varepsilon+|g_{x}|.$
    \STATE \textbf{If} adaptive: Update $p$ using gradient statistics (Eq.~\ref{eq:qstar})
    \STATE Compute attack $\delta_{p} \leftarrow \epsilon \cdot \text{sign}(g_{x})\cdot \left|\bar{g}_{x}/\| \bar{g}_{x} \|_{q}\right|^{q-1}$;
    \STATE Update $\theta$ with $\nabla_{\theta}\ell(x_{0}+\delta_{p},y_0)$ and optimizer;
    \UNTIL{Convergence criteria.}
    \STATE {\bfseries Output:} $l^p$-FGSM trained model $\theta$.
\end{algorithmic}
\end{algorithm}

\section{Experiments and Results}
\label{sec:exp}
Section~\ref{sec:theory} links CO to the concentration of the gradient (a drop of $\mathtt{PR_1}$) and predicts that adapting $p$ prevents it. We test this with fixed values of $p$ (Section~\ref{sec:fixedp}), the dynamics of adaptive $p$ (Section~\ref{sec:dynp}) and the comparison with other methods (Section~\ref{sec:valid}), then with long training and ImageNet (Sections~\ref{sec:long} and~\ref{sec:imagenet}).
The experimental settings are based on the protocols in Wong et al.~\cite{wong2020fast} and also De Jorge et al.~\cite{de2022make}. We use essentially 30 epochs for training SVHN~\cite{netzer2011reading} and CIFAR~\cite{krizhevsky2009learning} datasets and 15 epochs for ImageNet~\cite{deng2009imagenet}. We use PreAct-18~\cite{he2016identity} for SVHN, ResNet-50~\cite{he2016deep} for ImageNet and WRN-28-10~\cite{zagoruyko2016wide} for CIFAR datasets.

\subsection{Limitations of Fixed Norm Selection}
\label{sec:fixedp}
\begin{figure}[!t]
\centering
\includegraphics[width=0.98\linewidth]{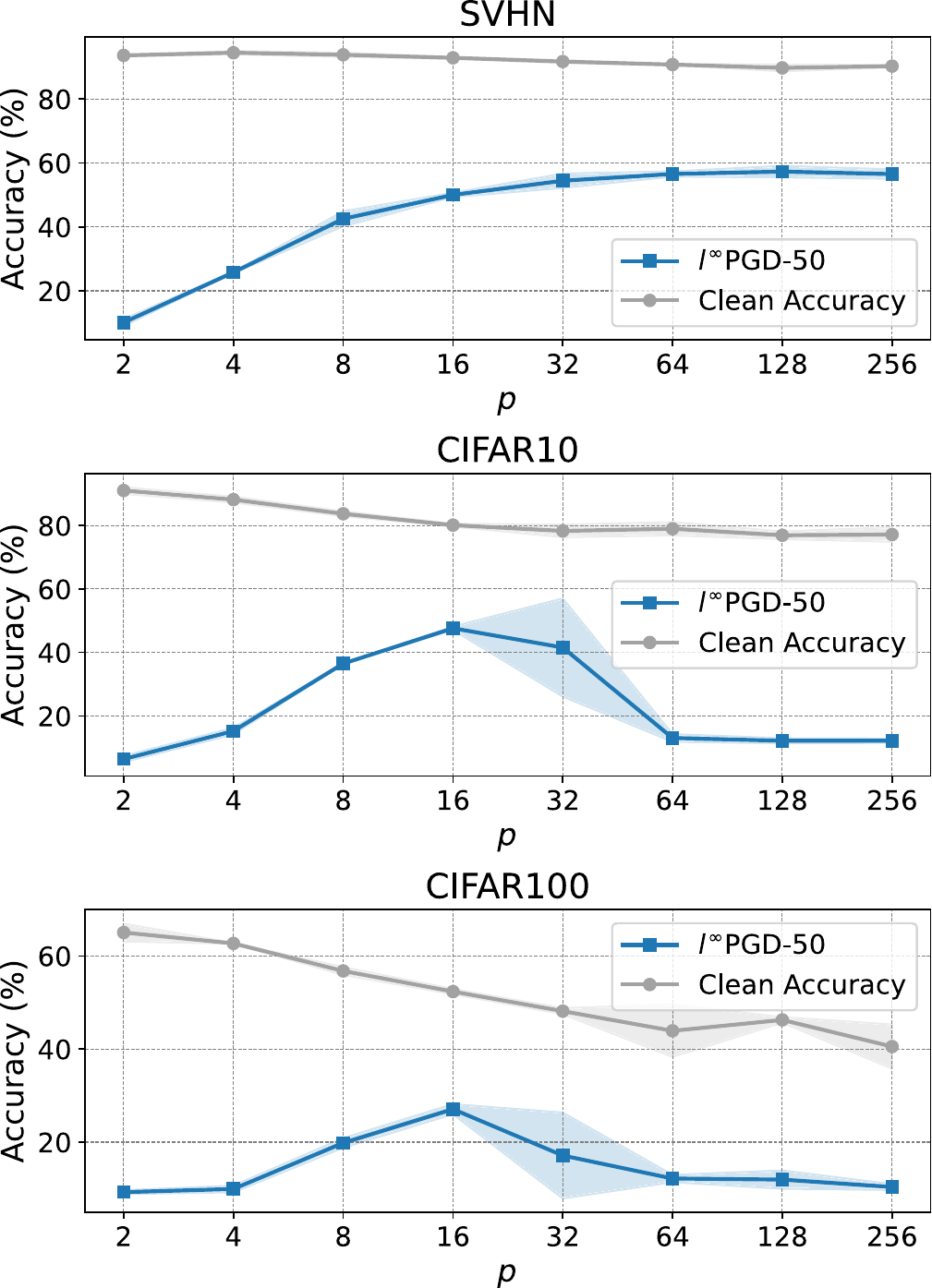}
\caption{Performance landscape of fixed $l^p$-FGSM across datasets ($\epsilon=8/255$). The results reveal a fundamental dilemma: lower $p$ values (e.g., $p=2,4$) ensure training stability but sacrifice adversarial robustness, achieving only 20-40\% accuracy against PGD-50. Higher values ($p \geq 32$) enhance robustness initially but inevitably trigger Catastrophic Overfitting (CO). Dataset complexity critically impacts this threshold: CIFAR-10 tolerates $p \approx 16$-$32$ before collapse, SVHN handles higher values, while CIFAR-100 exhibits heightened sensitivity with failure at $p \geq 16$. This dataset-dependent variability makes fixed norm selection impractical and motivates our adaptive approach. Training was done for 30 epochs with an ADAM~\cite{kingma2014adam} optimizer (lr=0.001, weight decay=0.0005) and a batch size of 128.}
\label{fig:fixed_p_limits}
\end{figure}

Before presenting our adaptive approach, we first show in more detail why fixed $p$ values fundamentally cannot solve the CO problem. Figure~\ref{fig:fixed_p_limits} reveals an inherent trade-off that varies dramatically across datasets and architectures.

The fundamental challenge is clear: no single $p$ value works universally. Lower values sacrifice too much robustness, while higher values risk Catastrophic Overfitting. This motivates our adaptive approach based on gradient geometry as detailed in Algorithm~\ref{alg:lp-fgsm}.

\subsection{Stability Under Adaptive Norm Selection}
\label{sec:dynp}

\begin{figure}[htbp]
\centering
\includegraphics[width=0.46\textwidth]{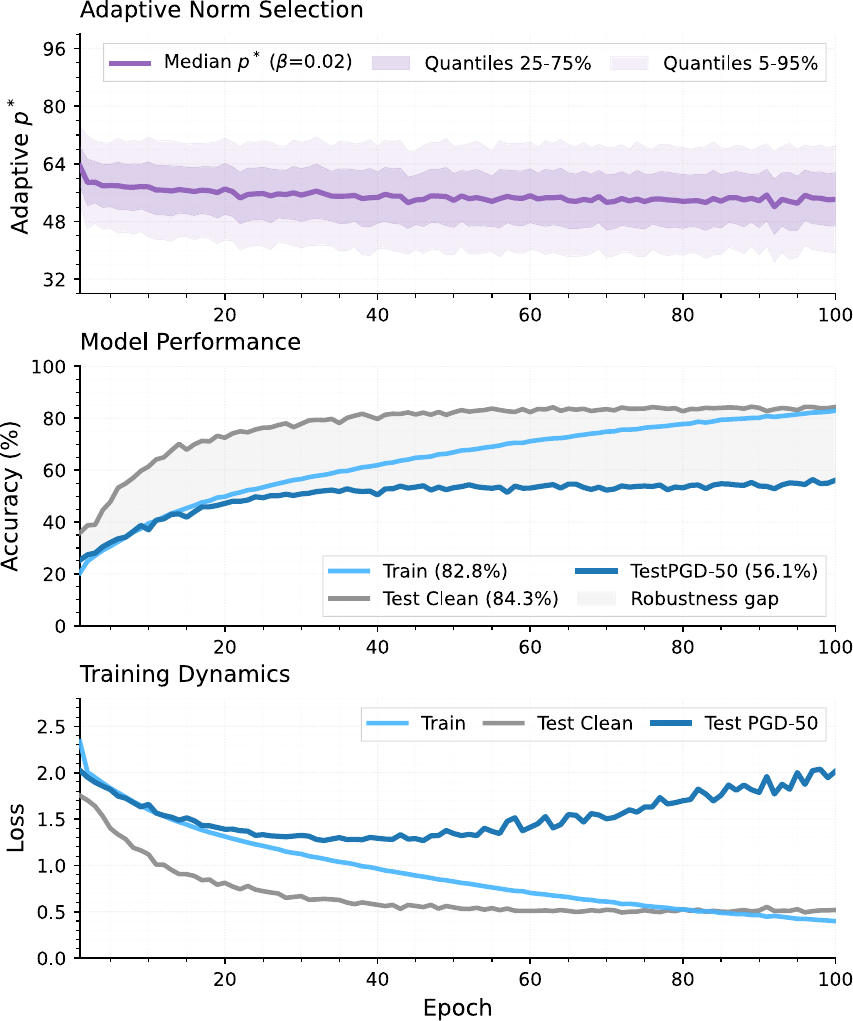}
\caption{Training dynamics of adaptive $l^p$-FGSM on CIFAR-10 with WideResNet-28-10 ($\beta=0.02$, $\epsilon=8/255$). ADAM~\cite{kingma2014adam} optimizer was used with constant learning rate of 0.001, a weight decay of 0.0005, and a batch size of 128. Despite continuous norm adaptation (top panel showing median and quartiles), training remains stable. The norm distribution maintains controlled variation within predictable bounds rather than erratic fluctuations. This translates to smooth accuracy progression (middle panel) without the oscillations or sudden drops characteristic of CO. The training loss (bottom panel) decreases steadily. While robust overfitting gradually emerges in later epochs, a known phenomenon in adversarial training, no catastrophic collapse occurs, validating that adaptive norm selection enhances rather than disrupts training stability.}
\label{fig:dynamic_p}
\end{figure}

A natural concern with adaptive methods is whether frequent norm adjustments might destabilize training. Figure~\ref{fig:dynamic_p} addresses this concern, demonstrating that our dynamic $p$ adjustment maintains stable convergence throughout training on CIFAR-10.
\vspace{0.1cm}

\subsection{Experimental Validation}
\label{sec:valid}

\begin{figure}[htbp]
\centering
\includegraphics[width=0.96\linewidth]{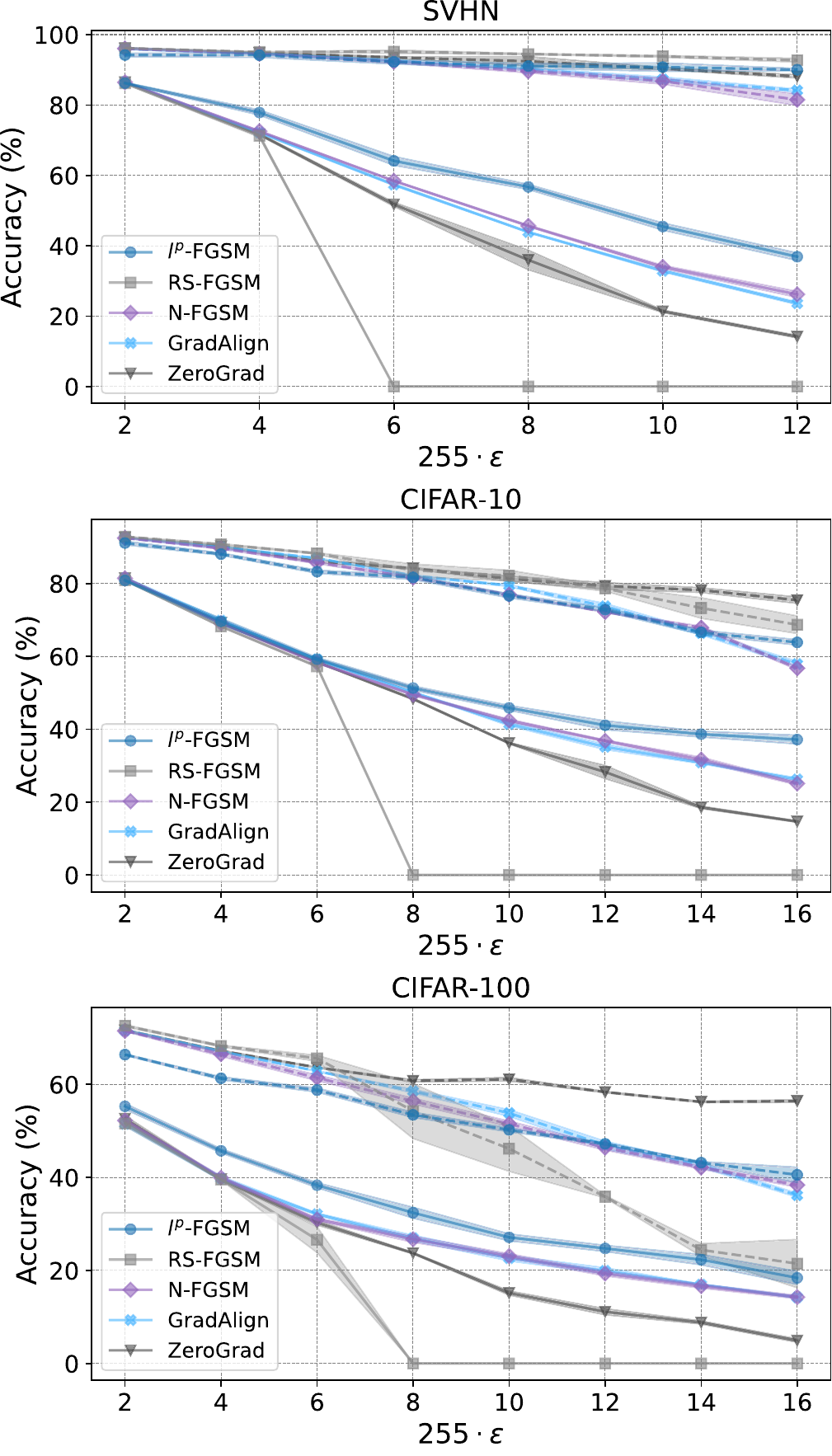}
\caption{Performance comparison of adaptive $l^p$-FGSM against established fast adversarial training methods. Evaluation uses PGD-50-10 attacks (50 steps, 10 restarts) for robust assessment. All methods trained for 30 epochs with single backpropagation, maintaining computational efficiency. Training employs SGD optimizer with cosine learning rate schedule (0.001-0.2), weight decay $5 \cdot 10^{-4}$, and dropout 0.1. Hyperparameter $\beta=0.01$ for CIFAR-10 and $\beta=0.1$ for CIFAR-100 (Eq.~\eqref{eq:hyper}), representing the only tunable parameter in our method. Extended 200-epoch stability analysis in Supplementary Materials Section VIII confirms long-term reliability.}
\label{fig:benchmark}
\end{figure}

Our empirical evaluation, summarized in Figure~\ref{fig:benchmark}, supports both our theoretical framework and practical effectiveness.

Our method successfully prevents catastrophic overfitting across all tested configurations, demonstrating consistent robust performance at $\epsilon=8/255$. These results are particularly notable given our minimal computational overhead, since we only add gradient norm calculations to standard FGSM. The performance is consistent with our core hypothesis: gradient concentration drives CO, and adaptive norm selection based on Participation Ratio and entropy gap provides an effective countermeasure.

Supplementary Materials Section~XII compares adaptive $l^p$-FGSM with recent single-step methods, including LIET~\cite{pan2025liet} and FGSM-PCO~\cite{wang2024preventing}, within an independent benchmark that applies one architecture and one protocol to all methods~\cite{pan2026fastat}. In that benchmark, adaptive $l^p$-FGSM trains at 1.02 times the cost of FGSM-AT; with a noisy start and a consistency term, it reaches 49.62\% AutoAttack accuracy on CIFAR-10 over three seeds, within 0.5 points of the best single-step methods.

\subsection{Long-term Stability Analysis}
\label{sec:long}
Practical deployment demands sustained performance beyond typical training schedules. To rigorously assess long-term stability, we conducted extended 200-epoch experiments on CIFAR-10 and CIFAR-100 (detailed in Supplementary Materials Section VIII). Our adaptive $l^p$-FGSM maintains robust performance throughout, never experiencing the sudden accuracy collapse characteristic of CO. While gradual robust overfitting occurs in later epochs, consistent with Rice et al.~\cite{rice2020overfitting}, the degradation is smooth and predictable. In contrast, FGSM and RS-FGSM already collapse within 30 epochs at larger $\epsilon$ (Supplementary Materials Section VII). This long-term stability, achieved without any additional regularization or noise injection, confirms our method's reliability for practical deployment scenarios where training schedules may vary.

\subsection{Scalability to High-Resolution Images}
\label{sec:imagenet}
To validate our approach beyond standard benchmarks, we evaluated adaptive $l^p$-FGSM on ImageNet-1k~\cite{ILSVRC15}, a dataset with 1000 classes and high-resolution images representative of real-world applications. We fine-tuned a pre-trained ResNet-50 model using ADAM optimizer (lr=$10^{-4}$, batch size 128) for 15 epochs, testing against PGD-50 attacks at multiple perturbation strengths. Our method uses $\beta=0.1$ and numerical stability parameter $\varepsilon=10^{-12}$.

\begin{table}[htbp]
\centering
\begin{small}
\renewcommand{\arraystretch}{1.2}
\caption{Robustness evaluation on ImageNet-1k against PGD-50-10 attacks. Baseline results from~\cite{de2022make}. All methods use ImageNet pre-trained weights with 15 epochs of adversarial fine-tuning. Results show clean accuracy (top) and robust accuracy (bottom) for each perturbation level.}
\begin{tabular}{|c|c|c|c|}
\hline
\multicolumn{4}{|c|}{\textbf{ImageNet-1k ResNet-50}} \\
\hline
\textbf{Method} & \textbf{$\epsilon = 2/255$} & \textbf{$\epsilon = 4/255$} & \textbf{$\epsilon = 6/255$} \\
\hline
FGSM & 54.72\% & 48.50\% & 48.55\% \\
     & \textbf{38.21\%} & 25.86\% & 0.08\% \\
\hline
RS-FGSM & \textbf{56.29\%} & \textbf{50.81\%} & 47.67\% \\
        & 36.86\% & 25.12\% & 16.49\% \\
\hline
N-FGSM & 54.39\% & 47.56\% & 47.70\% \\
       & 38.07\% & 26.28\% & 17.12\% \\
\hline
\textbf{$\boldsymbol{l^p}$-FGSM} & 53.18\% & 48.42\% & \textbf{48.61\%} \\
                                  & 37.94\% & \textbf{28.35\%} & \textbf{19.79\%} \\
\hline
\end{tabular}
\end{small}
\label{tab:imagenet_results}
\end{table}

Table~\ref{tab:imagenet_results} reveals a critical failure mode on real-world data: standard FGSM suffers complete catastrophic overfitting at $\epsilon=6/255$, achieving only 0.08\% robust accuracy, essentially random guessing. This dramatic collapse on ImageNet demonstrates that CO is not merely a toy dataset phenomenon but a fundamental vulnerability in fast adversarial training. Adaptive $l^p$-FGSM gives the highest robust accuracy at $\epsilon=4/255$ and $6/255$, a comparable one at $2/255$, and competitive clean accuracy. These results support our theoretical framework: by monitoring gradient concentration through PR$_1$ and entropy gap, we can prevent the geometric conditions that trigger CO, even in high-dimensional, complex datasets. This scalability from CIFAR-10 (3,072 dimensions) to ImageNet ($\sim$150,000 dimensions) supports the generality of our gradient concentration hypothesis and the practical applicability of adaptive $l^p$-FGSM for real-world robust training scenarios.

\section{Conclusion and Future Work}
\label{sec:conclusion}
Our study, inspired by the contrasting behaviors of $l^2$ and $l^\infty$ norms in adversarial training, provides new insights into the phenomenon of Catastrophic Overfitting (CO). By formulating $l^p$-norm bounded attacks as a fixed-point problem, we established connections to fundamental robustness metrics such as gradient alignment and normalized curvature through the Lipschitz constant.

Our key contribution revolves around identifying gradient concentration as a fundamental mechanism underlying CO, quantified through the novel application of Participation Ratio and entropy gap metrics. This theoretical framework enables real-time, adaptive norm selection during training, a computationally efficient solution requiring only gradient norm calculations without additional regularization, noise injection, or optimization overhead.

The development of $l^p$-FGSM demonstrated that uniformly reducing $p$ can delay the onset of CO but not entirely eliminate it, or, worse, undermine the adversarial robustness for too low values of $p$. This observation led us to a deeper geometric analysis, revealing how variations in $l^p$ norms influence effective dimensionality and impact the separation angle between $l^2$ and $l^\infty$ attacks. Our adaptive approach demonstrates competitive robust performance across standard benchmarks, while maintaining stability across extended 200-epoch training.

Our investigation revealed previously unexplored connections between attack geometry, entropy gap, and participation ratio, unifying concepts from machine learning, information theory, and quantum mechanics. This interdisciplinary perspective not only helps address the immediate CO problem but also provides a new theoretical lens for understanding adversarial vulnerability, complementing existing optimization-based approaches~\cite{andriushchenko2020understanding,zhang2022alleviating} and initialization strategies~\cite{wong2020fast, de2022make,jia2022prior} with geometric insights.

For immediate practical impact, our method offers: (1) a single hyperparameter $\beta$ that practitioners can easily tune, (2) negligible computational overhead suitable for resource-constrained environments, and (3) compatibility with existing training pipelines without architectural modifications. The ImageNet results indicate that the method scales to high-resolution data.
Future work will pursue three directions: (i) extending the fixed-point paradigm to multi-step training by optimally allocating the computational budget between inner-maximization iterations and fixed-point convergence steps; (ii) covering the full $l^p$ spectrum, including the non-smooth sub-quadratic regime $p\in[1,2]$; and (iii) transferring the Participation Ratio and entropy-gap metrics to related phase-transition phenomena such as neural collapse.

In summary, this work identifies gradient concentration as a key factor in adversarial robustness, providing both theoretical understanding and practical solutions to the CO problem. By showing that geometric properties of gradients influence training stability, we open new avenues for developing efficient and robust deep learning systems. Our approach's simplicity, theoretical grounding, and empirical effectiveness make it immediately applicable for practitioners while advancing the theoretical foundations of adversarial machine learning.

\section*{Acknowledgment}
This work was supported in part by the NYUAD Center for Interacting Urban Networks (CITIES), funded by Tamkeen under the NYUAD Research Institute Award CG001, and in part by the NYUAD Research Center on Stability, Instability, and Turbulence (SITE), funded by Tamkeen under the NYUAD Research Institute Award CG002. The views expressed in this article are those of the authors and do not reflect the opinions of CITIES, SITE, or their funding agencies.

We acknowledge the use of Grammarly and Claude.ai from Anthropic as writing aids throughout the preparation of this document.

\bibliographystyle{unsrt}
\bibliography{references}

\clearpage
\twocolumn[{\centering{\LARGE\sc Supplementary Materials}\par\vspace{1.2em}}]
\setcounter{section}{0}
\setcounter{figure}{0}
\setcounter{table}{0}
\setcounter{equation}{0}
\renewcommand{\thefigure}{S\arabic{figure}}
\renewcommand{\thetable}{S\arabic{table}}
\renewcommand{\theequation}{S\arabic{equation}}
\renewcommand{\theHfigure}{S\arabic{figure}}
\renewcommand{\theHtable}{S\arabic{table}}
\renewcommand{\theHequation}{S\arabic{equation}}
\renewcommand{\theHsection}{S\arabic{section}}
\section{Demonstration $l^2$ Optimal Attack}
\label{app:A}

\vspace{0.2cm}
\noindent\textbf{Proposition 1.} \textit{Consider a training sample $x_0$ with a non-null gradient. The optimal perturbation $\delta^{\star}$ within $B(\epsilon)$ exists and corresponds to the solution of a fixed-point problem $\delta^{\star}=F(\delta^{\star})$, where}
\begin{equation}
F(\delta) = \epsilon \frac{\nabla_{x}\ell(x_{0}+\delta)}{\|\nabla_{x}\ell(x_{0}+\delta)\|_2}
\end{equation}

\textit{The function $F$ exhibits Lipschitzian behavior around its origin, satisfying:}
\begin{equation}
\|F(\delta)-F(0)\| \leq 2\epsilon\frac{\|\nabla_{x}^{2}\ell\|}{\|\nabla_{x}\ell(x_{0})\|_{2}}\|\delta\|
\end{equation}

\textit{The fixed-point problem converges if it is contractive:}
\begin{equation}
    K=2\epsilon\frac{\|\nabla_{x}^{2}\ell\|}{\|\nabla_{x}\ell(x_{0})\|_2}<1
\end{equation}
\vspace{0.2cm}

\begin{proof}
Assuming that the Hessian of the loss function, $\nabla^{2}_x\ell$, is positive ($\nabla^{2}_x\ell\succeq 0$) on $B_2(\epsilon)$, the loss is convex on this compact ball and attains its maximum on the boundary. The constrained maximization uses the Lagrangian:
\begin{equation}
\mathcal{L}(\delta,\lambda)=\ell(x_{0}+\delta)-\frac{\lambda}{2}(\delta^{T}\delta-\epsilon^{2})
\end{equation}

The derivatives yield the following equations:
\begin{equation}
\begin{cases}
\frac{\partial}{\partial\delta}\mathcal{L}=\nabla_{x}\ell(x_{0}+\delta)-\lambda\delta=0\\
\frac{\partial}{\partial\lambda}\mathcal{L}=-\frac{1}{2}(\delta^{T}\delta-\epsilon^{2})=0
\end{cases}
\end{equation}

Since the maximum exists on the boundary, the constraint $\delta^T\delta=\epsilon^2$ is activated; hence the Lagrange multiplier $\lambda$ is non-null. The gradient at $x_0+\delta$ cannot be null (minimum otherwise), therefore $\|\nabla_{x}\ell(x_{0}+\delta)\|>0$.

Solving the two Lagrangian equations yields:
\begin{equation}
\delta=\pm\epsilon\frac{\nabla_{x}\ell(x_{0}+\delta)}{\|\nabla_{x}\ell(x_{0}+\delta)\|}
\end{equation}

Given the positive Hessian assumption, moving along the gradient (equivalent to choosing the positive sign) results in a greater change in the loss function $\ell$. Consequently:
\begin{equation}
\delta=\epsilon\frac{\nabla_{x}\ell(x_{0}+\delta)}{\|\nabla_{x}\ell(x_{0}+\delta)\|}
\end{equation}

The maximum $\delta^\star$ is the solution to a fixed-point problem. The existence and uniqueness of the solution $\delta^\star$ is guaranteed if $F(\delta)$ is contractive, i.e., Lipschitz continuous with a Lipschitz constant $K<1$.

To demonstrate this Lipschitz continuity, we consider:
\begin{align}
\|F(\delta)-F(0)\| =\epsilon\Big\|\frac{\nabla_{x}\ell(x_{0}+\delta)}{\|\nabla_{x}\ell(x_{0}+\delta)\|} \nonumber \\
-\frac{\nabla_{x}\ell(x_{0})}{\|\nabla_{x}\ell(x_{0})\|}\Big\|
\end{align}

By introducing a cross term and using the triangle inequality:
\begin{align}
\|F(\delta)-F(0)\| \leq \epsilon\Big\|\frac{\nabla_{x}\ell(x_{0})}{\|\nabla_{x}\ell(x_{0})\|} \nonumber \\
-\frac{\nabla_{x}\ell(x_{0}+\delta)}{\|\nabla_{x}\ell(x_{0})\|}\Big\| \nonumber \\
+ \epsilon\Big\|\frac{\nabla_{x}\ell(x_{0}+\delta)}{\|\nabla_{x}\ell(x_{0})\|} \nonumber \\
-\frac{\nabla_{x}\ell(x_{0}+\delta)}{\|\nabla_{x}\ell(x_{0}+\delta)\|}\Big\|
\end{align}

The first term can be bounded:
\begin{align}
\|F(\delta)-F(0)\| \leq\epsilon\frac{\|\nabla_{x}^{2}\ell(x_{0})\|\|\delta\|}{\|\nabla_{x}\ell(x_{0})\|} \nonumber \\
+\epsilon\|\nabla_{x}\ell(x_{0}+\delta)\|\Big|\frac{1}{\|\nabla_{x}\ell(x_{0}+\delta)\|} \nonumber \\
-\frac{1}{\|\nabla_{x}\ell(x_{0})\|}\Big|
\end{align}

After unifying the denominator:
\begin{align}
\|F(\delta)-F(0)\| \leq\epsilon\frac{\|\nabla_{x}^{2}\ell\|\|\delta\|}{\|\nabla_{x}\ell(x_{0})\|} \nonumber \\
+\frac{\epsilon}{\|\nabla_{x}\ell(x_{0})\|}\Big|\|\nabla_{x}\ell(x_{0}+\delta)\| \nonumber \\
-\|\nabla_{x}\ell(x_{0})\|\Big|
\end{align}

Using the triangle inequality again:
\begin{align}
\Big|\|\nabla_{x}\ell(x_{0}+\delta)\|-\|\nabla_{x}\ell(x_{0})\|\Big| \nonumber \\
\leq \|\nabla_{x}\ell(x_{0}+\delta)-\nabla_{x}\ell(x_{0})\| \nonumber \\
\leq\|\nabla_{x}^{2}\ell\|\|\delta\|
\end{align}

This leads to:
\begin{equation}
\|F(\delta)-F(0)\| \leq 2\epsilon\frac{\|\nabla_{x}^{2}\ell(x_{0})\|\|\delta\|}{\|\nabla_{x}\ell(x_{0})\|}
\end{equation}

The Lipschitz constant is:
\begin{equation}
K=2\epsilon\cdot\frac{\|\nabla_{x}^{2}\ell(x_{0})\|}{\|\nabla_{x}\ell(x_{0})\|}
\end{equation}

This bound compares $F(\delta)$ with $F(0)$. A contraction needs the same type of bound for any two points, which follows from the same quantities. Let $\|\nabla_{x}^{2}\ell\|$ denote the supremum of the spectral norm of the Hessian over $B_2(\epsilon)$. Then $\|\nabla_{x}\ell(x_{0}+\delta)-\nabla_{x}\ell(x_{0}+\delta')\|\leq\|\nabla_{x}^{2}\ell\|\|\delta-\delta'\|$ and, for $\delta\in B_2(\epsilon)$, $\|\nabla_{x}\ell(x_{0}+\delta)\|\geq\|\nabla_{x}\ell(x_{0})\|-\epsilon\|\nabla_{x}^{2}\ell\|=\|\nabla_{x}\ell(x_{0})\|(1-K/2)$. In an inner-product space, $\|a/\|a\|-b/\|b\|\|\leq2\|a-b\|/(\|a\|+\|b\|)$ for non-null $a$ and $b$ (Dunkl--Williams inequality). Hence, for $\delta,\delta'\in B_2(\epsilon)$,
\[
\|F(\delta)-F(\delta')\|\leq\frac{\epsilon\|\nabla_{x}^{2}\ell\|\|\delta-\delta'\|}{\|\nabla_{x}\ell(x_{0})\|(1-K/2)}=\frac{K}{2-K}\|\delta-\delta'\|.
\]
If $K<1$, then $K/(2-K)<1$. Since $F$ maps $B_2(\epsilon)$ into its boundary, $F$ is a contraction of $B_2(\epsilon)$, and by the Banach fixed-point theorem the fixed point is unique and the iteration $\delta^{(k+1)}=F(\delta^{(k)})$ converges to it from any starting point in $B_2(\epsilon)$, including $\delta^{(0)}=0$.
\end{proof}

\section{Demonstration $l^p$ Optimal Attack}
\label{app:B}

\vspace{0.2cm}
\noindent\textbf{Proposition 2.} \textit{For a training sample $x_0$ exhibiting a non-null gradient and a constraint within $B_p(\epsilon)$, the optimal perturbation, denoted as $\delta^\star$, exists and corresponds to the solution of a fixed-point problem: $\delta^{\star}=F_p(\delta^{\star})$. Specifically, we have:}
\begin{equation}
F_{p}(\delta)=\epsilon\text{sign}(\nabla_{x}\ell(x_{0}+\delta))\left|\frac{\nabla_{x}\ell(x_{0}+\delta)}{\|\nabla_{x}\ell(x_{0}+\delta)\|_{q}}\right|^{q-1}
\end{equation}

\textit{where the $l^q$ norm serves as the dual to $l^p$, i.e., $\frac{1}{p}+\frac{1}{q}=1$. The absolute value and multiplication operations are element-wise.}
\vspace{0.2cm}

\begin{proof}
Assuming the same hypotheses as in Section I, a maximum exists on the boundary of the $B_p$ ball. We formulate the Lagrangian with the $l^p$ equality constraint:
\begin{equation}
\mathcal{L}_{p}(\delta,\lambda)=\ell(x_{0}+\delta)-\lambda(\|\delta\|_{p}-\epsilon)
\end{equation}

The $l^p$ norm is given by:
\begin{equation}
\|\delta\|_{p}=\left(\sum_{i}|\delta_{i}|^{p}\right)^{\frac{1}{p}}
\end{equation}

Hence, its derivative is:
\begin{equation}
\frac{\partial}{\partial\delta}\|\delta\|_{p}=\text{sign}(\delta)\left(\frac{|\delta|}{\|\delta\|_{p}}\right)^{p-1}
\end{equation}

The derivatives of the Lagrangian are:
\begin{equation}
\begin{cases}
\frac{\partial}{\partial\delta}\mathcal{L}_{p}=\nabla_{x}\ell(x_{0}+\delta)-\lambda\text{sign}(\delta)\left(\frac{|\delta|}{\|\delta\|_{p}}\right)^{p-1}=0\\
\frac{\partial}{\partial\lambda}\mathcal{L}_{p}=-(\|\delta\|_{p}-\epsilon)=0
\end{cases}
\end{equation}

Using the dual norm $l^q$ defined with $\frac{1}{p}+\frac{1}{q}=1\rightarrow q=\frac{p}{p-1}$, we can characterize $\lambda$ as:
\begin{align}
\|\nabla_{x}\ell(x_{0}+\delta)\|_{q} &= \frac{|\lambda|}{\|\delta\|_{p}^{p-1}}(\|\delta\|_{p}^{p})^{\frac{1}{q}} \nonumber \\
&= |\lambda|
\end{align}

Substituting into the first derivative of the Lagrangian:
\begin{align}
\nabla_{x}\ell(x_{0}+\delta) &= \pm\|\nabla_{x}\ell(x_{0}+\delta)\|_{q} \nonumber \\
&\times \text{sign}(\delta)\left(\frac{|\delta|}{\|\delta\|_{p}}\right)^{p-1}
\end{align}

From this, $\delta$ and $\nabla_{x}\ell(x_{0}+\delta)$ have the same sign up to a multiplicative coefficient (i.e., $\pm$):
\begin{align}
\frac{\nabla_{x}\ell(x_{0}+\delta)}{\|\nabla_{x}\ell(x_{0}+\delta)\|_{q}} = \pm\left|\frac{\delta}{\|\delta\|_{p}}\right|^{p-1}\text{sign}(\delta)
\end{align}

Extracting $\delta$ and using $\|\delta\|_{p}=\epsilon$ yields:
\begin{align}
\delta = \pm\epsilon\text{sign}(\nabla_{x}\ell(x_{0}+\delta)) \nonumber \\
\times \left(\frac{|\nabla_{x}\ell(x_{0}+\delta)|}{\|\nabla_{x}\ell(x_{0}+\delta)\|_{q}}\right)^{\frac{1}{p-1}}
\end{align}

The solution with the negative sign would yield a locally decreasing loss function, so we take the positive solution. The Lagrange multiplier for maximization is positive:
\begin{equation}
\lambda=\|\nabla_{x}\ell(x_{0}+\delta)\|_{q}
\end{equation}

Using $p=\frac{q}{q-1}\rightarrow p-1=\frac{1}{q-1}$, we get the final result:
\begin{align}
\delta = \epsilon\text{sign}(\nabla_{x}\ell(x_{0}+\delta)) \nonumber \\
\times \left(\frac{|\nabla_{x}\ell(x_{0}+\delta)|}{\|\nabla_{x}\ell(x_{0}+\delta)\|_{q}}\right)^{q-1}
\end{align}
\end{proof}

\section{Lipschitzness of the $l^p$ Fixed-Point Problem}
\label{app:C}

We assume: $\exists\,m>0:\forall\delta\in\partial B_{p}(\epsilon),|\nabla_{x}\ell(x_{0}+\delta)_{i}|>m$, and proceed to demonstrate Lipschitzness of the function $F_p(\delta)$ verifying the fixed point, defined as:
\begin{equation}
F_p(\delta) = \epsilon\,\text{sign}(\nabla_x\ell(x_0+\delta))\left|\frac{\nabla_x\ell(x_0+\delta)}{\|\nabla_x\ell(x_0+\delta)\|_q}\right|^{q-1}
\end{equation}

The sign function can be circumvented by using ``one power'' of the absolute value of the gradient:
\begin{align}
F_p(\delta) &= \epsilon\frac{\nabla_{x}\ell(x_{0}+\delta)}{\|\nabla_{x}\ell(x_{0}+\delta)\|_q} \nonumber \\
&\times \left(\frac{|\nabla_{x}\ell(x_{0}+\delta)|}{\|\nabla_{x}\ell(x_{0}+\delta)\|_{q}}\right)^{q-2}
\end{align}

The term $q-2$ is negative, which is permissible since we assumed a lower limit $m$ for gradient values. Our objective is to prove that $F_p(\delta)$ is Lipschitz continuous around $\delta = 0$.

First, let's define:
\begin{equation}
f_q(\delta) = \frac{\nabla_x\ell(x_0+\delta)}{\|\nabla_x\ell(x_0+\delta)\|_q}
\end{equation}

We have:
\begin{equation}
F_{p}(\delta) = \epsilon f_{q}(\delta)|f_{q}(\delta)|^{q-2}
\end{equation}

Similar to Supplementary Materials Section I, by introducing a cross term we can show that $f$ and $|f|$ are Lipschitz continuous, with a constant $K_f$ such that:
\begin{equation}
|f_q(\delta) - f_q(0)| \leq K_f\|\delta\|
\end{equation}

The same steps are applied as follows:
\begin{align}
\||f_{q}(\delta)| - |f_{q}(0)|\| &\leq \|f_{q}(\delta) - f_{q}(0)\| \nonumber \\
&\leq \Big\|\frac{\nabla_{x}\ell(x_{0}+\delta)}{\|\nabla_{x}\ell(x_{0}+\delta)\|_{q}} \nonumber \\
&- \frac{\nabla_{x}\ell(x_{0})}{\|\nabla_{x}\ell(x_{0})\|_{q}}\Big\|
\end{align}

By further manipulation and using the triangle inequality:
\begin{align}
\||f_{q}(\delta)| - |f_{q}(0)|\| &\leq \Big\|\frac{\nabla_{x}\ell(x_{0}+\delta)}{\|\nabla_{x}\ell(x_{0}+\delta)\|_{q}} \nonumber \\
&- \frac{\nabla_{x}\ell(x_{0}+\delta)}{\|\nabla_{x}\ell(x_{0})\|_{q}}\Big\| \nonumber \\
&+ \Big\|\frac{\nabla_{x}\ell(x_{0})}{\|\nabla_{x}\ell(x_{0})\|_{q}} \nonumber \\
&- \frac{\nabla_{x}\ell(x_{0}+\delta)}{\|\nabla_{x}\ell(x_{0})\|_{q}}\Big\|
\end{align}

This leads to:
\begin{align}
\||f_{q}(\delta)| - |f_{q}(0)|\| &\leq \left(1 + \frac{\|\nabla_{x}\ell(x_{0}+\delta)\|}{\|\nabla_{x}\ell(x_{0}+\delta)\|_{q}}\right) \nonumber \\
&\times \frac{\|\nabla_{x}^{2}\ell(x_{0})\|}{\|\nabla_{x}\ell(x_{0})\|_{q}}\|\delta\|
\end{align}
In a finite-dimensional vector space, all norms are equivalent:
\begin{equation}
\exists\,C\geq 0,\,\frac{\|\nabla_{x}\ell(x_{0}+\delta)\|}{\|\nabla_{x}\ell(x_{0}+\delta)\|_{q}}\leq C
\end{equation}
Next, examining $|x|^{q-2}$ on the interval $[m,+\infty)$ with $q-2$ negative:
\begin{equation}
\forall(x,y)\in[m,+\infty),\,||x|^{q-2}-|y|^{q-2}|\leq(2-q)m^{q-3}|x-y|
\end{equation}
Using these results for the local Lipschitz continuity of $F_p$:
\begin{align}
\frac{1}{\epsilon}\|F_{p}(\delta)-F_{p}(0)\| &= \|f_{q}(\delta)|f_{q}(\delta)|^{q-2} \nonumber \\
&- f_{q}(0)|f_{q}(0)|^{q-2}\|
\end{align}

Through a series of bounds:
\begin{align}
\frac{1}{\epsilon}\|F_{p}(\delta)-F_{p}(0)\| &\leq \|f_{q}(\delta)|f_{q}(\delta)|^{q-2} \nonumber \\
&- f_{q}(\delta)|f_{q}(0)|^{q-2}\| \nonumber \\
&+ \|f_{q}(\delta)|f_{q}(0)|^{q-2} \nonumber \\
&- f_{q}(0)|f_{q}(0)|^{q-2}\|
\end{align}

Further simplifying:
\begin{align}
\frac{1}{\epsilon}\|F_{p}(\delta)-F_{p}(0)\| &\leq \frac{\|\nabla_{x}\ell(x_{0}+\delta)\|}{\|\nabla_{x}\ell(x_{0}+\delta)\|_{q}} \nonumber \\
&\times (2-q)m^{q-3}|f_{q}(\delta)-f_{q}(0)| \nonumber \\
&+ \left|\frac{\nabla_{x}\ell(x_{0})}{\|\nabla_{x}\ell(x_{0})\|_{q}}\right|^{q-2} \nonumber \\
&\times \|f_{q}(\delta)-f_{q}(0)\|
\end{align}

This yields:
\begin{align}
\|F_{p}(\delta)-F_{p}(0)\| &\leq K(p,m)\epsilon \nonumber \\
&\times \frac{\|\nabla_{x}^{2}\ell(x_{0})\|}{\|\nabla_{x}\ell(x_{0})\|_{q}}\|\delta\|
\end{align}

where:
\begin{align}
K(p,m) &= \left(C(2-q)m^{q-3} \right. \nonumber \\
&\left. + \left(\frac{m}{\|\nabla_{x}\ell(x_{0})\|_{q}}\right)^{q-2}\right)(1+C)
\end{align}

\section{Proof of Lemma 1}
\label{app:D}

\noindent\textbf{Lemma 1 (Noise-Induced Alignment).} \textit{Let $g\in\mathbb{R}^d\setminus\{0\}$, $M>0$, $\eta\sim\mathcal{U}[-M,M]^d$, and
\[
\kappa_M=\frac{1}{dM^2}\sum_{i=1}^{d}(M-|g_i|)_+^2,\qquad \nu_M=\frac{M\|g\|_1}{\|g\|_2^2}.
\]
Then $\mathbb{E}\|g+\eta\|_1=\|g\|_1+\frac{dM}{2}\kappa_M$ and $\mathbb{E}\|g+\eta\|_2^2=\|g\|_2^2+\frac{dM^2}{3}$. If $\kappa_M>\nu_M/3$, then
\begin{equation}
\mathbb{P}\big[\mathtt{PR_1}(g+\eta)\ge\mathtt{PR_1}(g)\big]\ge1-\exp\!\left(-\frac{d\,(\kappa_M-\nu_M/3)^2}{8\,(\sin\theta_{2,\infty}+\nu_M)^2}\right),
\label{eq:S-lemma1}
\end{equation}
where $\cos\theta_{2,\infty}=\|g\|_1/(\sqrt{d}\,\|g\|_2)=\sqrt{\mathtt{PR_1}(g)/d}$ as in the main text.}

\noindent\textit{Notation.} Write $t_i=|g_i|$ and $\bar m=\|g\|_2^2/\|g\|_1$, so that $\nu_M=M/\bar m$ and $\|g\|_1/\bar m=\|g\|_2^2/\bar m^2=\mathtt{PR_1}(g)$. Since $-\eta_i$ has the law of $\eta_i$, the vector $(|g_i+\eta_i|)_i$ has the law of $(|t_i+\eta_i|)_i$. Both norms depend only on this vector, so we may assume $g_i=t_i\ge0$.

\begin{proof}
\textit{1) Expected norms.} For one coordinate,
$\mathbb{E}|t_i+\eta_i|=\frac{1}{2M}\int_{-M}^{M}|t_i+u|\,du$. If $t_i\ge M$, the sign does not change on the interval and the average equals $t_i$. If $t_i<M$, splitting at $u=-t_i$ gives $\big((M-t_i)^2+(M+t_i)^2\big)/(4M)=(t_i^2+M^2)/(2M)$. Both cases read
\begin{equation}
\mathbb{E}|t_i+\eta_i|=t_i+\frac{(M-t_i)_+^2}{2M}.
\label{eq:S-fold}
\end{equation}
For $t_i<M$, the gain equals twice the probability $(M-t_i)/(2M)$ that the sign flips, times the mean overshoot $(M-t_i)/2$ past zero. Both factors are linear in $M-t_i$, hence the square. Summing \eqref{eq:S-fold} over $i$ gives $\mathbb{E}\|g+\eta\|_1=\|g\|_1+\frac{dM}{2}\kappa_M$. Since $0\le(M-t_i)_+\le M$, $\kappa_M\in[0,1]$, and $\kappa_M=0$ exactly when $M\le\min_i t_i$. Finally, $\mathbb{E}\eta_i=0$ and $\mathbb{E}\eta_i^2=M^2/3$ give $\mathbb{E}\|g+\eta\|_2^2=\|g\|_2^2+dM^2/3$.

\textit{2) A deterministic sufficient condition.} For every $x\in\mathbb{R}^d$, the square root lies below its tangent at $\|g\|_2^2$, so $\|x\|_2\le\|g\|_2+(\|x\|_2^2-\|g\|_2^2)/(2\|g\|_2)$. Multiplying by $\|g\|_1/\|g\|_2$,
\begin{equation*}
\frac{\|g\|_1}{\|g\|_2}\,\|x\|_2\le\|g\|_1+\frac{\|x\|_2^2-\|g\|_2^2}{2\bar m}.
\end{equation*}
Hence, with
\begin{equation}
W(x)=\|x\|_1-\|g\|_1-\frac{\|x\|_2^2-\|g\|_2^2}{2\bar m},
\label{eq:S-W}
\end{equation}
$W(x)\ge0$ implies $\|x\|_1/\|x\|_2\ge\|g\|_1/\|g\|_2$, that is, $\mathtt{PR_1}(x)\ge\mathtt{PR_1}(g)$. ($W(0)=-\|g\|_1/2<0$, so $x\neq0$ whenever $W(x)\ge0$.)

\textit{3) Mean.} By step 1,
\begin{equation}
\mathbb{E}\,W(g+\eta)=\frac{dM}{2}\kappa_M-\frac{dM^2}{6\bar m}=\frac{dM}{2}\Big(\kappa_M-\frac{\nu_M}{3}\Big).
\label{eq:S-mean}
\end{equation}

\textit{4) Concentration.} $W(g+\eta)=\sum_{i=1}^{d}\psi_i(\eta_i)$ is a sum of independent terms, with $\psi_i(u)=|t_i+u|-t_i-\big((t_i+u)^2-t_i^2\big)/(2\bar m)$. Where $t_i+u\neq0$, $\psi_i'(u)=\operatorname{sign}(t_i+u)-(t_i+u)/\bar m$. For $|u|\le M$: if $t_i+u>0$, then $|\psi_i'(u)|=|1-(t_i+u)/\bar m|\le|1-t_i/\bar m|+\nu_M$; if $t_i+u<0$, which requires $t_i<M$, then $|\psi_i'(u)|=|1-|t_i+u|/\bar m|\le1+(M-t_i)/\bar m\le|1-t_i/\bar m|+\nu_M$. Hence $\psi_i$ is $L_i$-Lipschitz on $[-M,M]$ with $L_i=|1-t_i/\bar m|+\nu_M$, and $\psi_i(\eta_i)$ lies in an interval of length at most $2ML_i$. When $\kappa_M>\nu_M/3$, \eqref{eq:S-mean} gives $\mathbb{E}\,W(g+\eta)>0$, and Hoeffding's inequality~\cite{hoeffding1963probability} gives
\begin{equation}
\mathbb{P}\big[W(g+\eta)<0\big]\le\exp\!\left(-\frac{\big(\mathbb{E}\,W(g+\eta)\big)^2}{2M^2\sum_{i}L_i^2}\right).
\label{eq:S-hoeffding}
\end{equation}
By Minkowski's inequality,
\begin{equation*}
\Big(\sum_{i=1}^{d}L_i^2\Big)^{1/2}\le\Big(\sum_{i=1}^{d}\Big(1-\frac{t_i}{\bar m}\Big)^2\Big)^{1/2}+\sqrt{d}\,\nu_M,
\end{equation*}
and
\begin{equation*}
\sum_{i=1}^{d}\Big(1-\frac{t_i}{\bar m}\Big)^2=d-\frac{2\|g\|_1}{\bar m}+\frac{\|g\|_2^2}{\bar m^2}=d-\mathtt{PR_1}(g)=d\sin^2\theta_{2,\infty}.
\end{equation*}
Thus $\sum_i L_i^2\le d\,(\sin\theta_{2,\infty}+\nu_M)^2$. Inserting this bound and \eqref{eq:S-mean} into \eqref{eq:S-hoeffding}, and using step 2, $\mathbb{P}[\mathtt{PR_1}(g+\eta)\ge\mathtt{PR_1}(g)]\ge\mathbb{P}[W(g+\eta)\ge0]=1-\mathbb{P}[W(g+\eta)<0]$ yields \eqref{eq:S-lemma1}.
\end{proof}

\noindent\textit{Remarks.} (i) \emph{Weak noise.} If $M<\min_i t_i$, then $\kappa_M=0$, no coordinate changes sign, and $\mathtt{PR_1}$ is smooth on the support of $g+\eta$. With $A=\|x\|_1$ and $Q=\|x\|_2^2$, $\partial_i^2(A^2/Q)=2/Q-8A|x_i|/Q^2-2A^2/Q^2+8A^2x_i^2/Q^3$ away from the coordinate hyperplanes, so
\begin{equation*}
\sum_{i=1}^{d}\partial_i^2\,\mathtt{PR_1}(x)=\frac{2d}{\|x\|_2^2}\big(1-\mathtt{PR_1}(x)\big).
\end{equation*} In a Taylor expansion around $g$, the first- and third-order terms and the mixed second-order terms vanish in expectation, which gives
\begin{equation}
\mathbb{E}\,\mathtt{PR_1}(g+\eta)=\mathtt{PR_1}(g)-\frac{dM^2}{3\|g\|_2^2}\big(\mathtt{PR_1}(g)-1\big)+\mathcal{O}(M^4).
\label{eq:S-weak}
\end{equation}
For $d\ge2$, $\mathtt{PR_1}(g)>1$, so noise weaker than every coordinate lowers the expected $\mathtt{PR_1}$ to leading order in $M$.

(ii) \emph{Many small coordinates.} If a fraction $f$ of the coordinates satisfies $|g_i|\le\varepsilon$, then $\kappa_{3\varepsilon}\ge f(1-\tfrac13)^2=\tfrac49f$, while $\nu_{3\varepsilon}/3=\varepsilon\|g\|_1/\|g\|_2^2$. The lemma therefore applies at noise level $M=3\varepsilon$ as soon as $f>\frac{9}{4}\,\varepsilon\|g\|_1/\|g\|_2^2$. This condition is mild for concentrated gradients, whose $l^1$-weighted mean magnitude $\|g\|_2^2/\|g\|_1$ lies far above most coordinates.

(iii) \emph{Sharper threshold.} The condition $\kappa_M>\nu_M/3$ comes from the tangent at $\|g\|_2^2$ in step 2. The same argument with the tangent at $\mathbb{E}\|g+\eta\|_2^2$ gives a bound of the same type, with modified constants, under the weaker condition $\big(\mathbb{E}\|g+\eta\|_1\big)^2>\mathtt{PR_1}(g)\,\mathbb{E}\|g+\eta\|_2^2$, that is, $\kappa_M>(2\nu_M/3)/\big(1+\sqrt{1+\rho_M}\big)$ with $\rho_M=dM^2/(3\|g\|_2^2)$. This threshold reduces to $\nu_M/3$ as the relative noise energy $\rho_M$ tends to zero.

(iv) \emph{Scope.} The lemma concerns additive noise in gradient space. Input noise changes the gradient through the network and need not produce independent uniform gradient noise.

\section{Proof of Monotonicity of Angular Separation}
\label{app:E}
\vspace{0.3cm}

\noindent\textbf{Lemma 2 (Monotonicity of Angular Separation).} \textit{For any gradient $\nabla_x \ell$ and $2 \leq p \leq \infty$, the cosine similarity between $l_2$ and $l_p$ perturbations satisfies:}
\begin{equation}
\cos(\theta_{2,p}) \geq \cos(\theta_{2,\infty}) = \sqrt{\frac{\mathtt{PR}_1}{d}}
\end{equation}

\vspace{0.3cm}
\begin{proof}

\vspace{0.2cm}
\textbf{Step 1: Express $\cos(\theta_{2,p})$ in normalized form.}

\vspace{0.2cm}
Let $q = \frac{p}{p-1}$ be the dual exponent of $p$; hence $2 \leq p \leq \infty$ implies $1 \leq q \leq 2$. Recall that:
\begin{align}
\delta_p &= \epsilon\,\text{sign}(\nabla_{x}\ell(x_{0}))\left|\frac{\nabla_{x}\ell(x_{0})}{\|\nabla_{x}\ell(x_{0})\|_{q}}\right|^{q-1} \\
\delta_\infty &= \epsilon\,\text{sign}(\nabla_{x}\ell(x_{0}))
\end{align}

The cosine similarity between the two perturbations is:
\begin{equation}
\cos(\theta_{2,p}) = \frac{\langle\delta_{2},\delta_{p}\rangle}{\|\delta_{2}\|_{2}\|\delta_{p}\|_{2}}
\end{equation}

After computing the inner product and norms, this yields:
\begin{align}
\cos(\theta_{2,p}) &= \frac{\|\nabla_x \ell\|_q^q}{\|\nabla_x \ell\|_2 \,\|\nabla_x \ell\|_{2(q-1)}^{q-1}}
\end{align}

We introduce the normalized vector:
\begin{equation}
g = \frac{\nabla_x \ell}{\|\nabla_x \ell\|_2}
\end{equation}

Note that $\|g\|_2 = 1$, and each coordinate satisfies $|g_i| \leq 1$. Using $g$:
\begin{align}
\|\nabla_x \ell\|_q &= \|\nabla_x \ell\|_2 \,\|g\|_q\\
\|\nabla_x \ell\|_q^q &= \|\nabla_x \ell\|_2^q \,\|g\|_q^q\\
\|\nabla_x \ell\|_{2(q-1)}^{q-1} &= \|\nabla_x \ell\|_2^{q-1} \,\|g\|_{2(q-1)}^{q-1}
\end{align}

Substituting these into our expression for $\cos(\theta_{2,p})$:
\begin{align}
\cos(\theta_{2,p}) &= \frac{\|\nabla_x \ell\|_2^q \,\|g\|_q^q}{\|\nabla_x \ell\|_2 \,\|\nabla_x \ell\|_2^{q-1}\,\|g\|_{2(q-1)}^{q-1}} \\[0.3em]
&= \frac{\|g\|_q^q}{\|g\|_{2(q-1)}^{q-1}} = \frac{\|g\|_q^q}{\sqrt{\|g\|_{2(q-1)}^{2(q-1)}}}
\end{align}

\vspace{0.4cm}
\textbf{Step 2: Show monotonicity via logarithmic derivative.}

\vspace{0.2cm}
Define:
\begin{equation}
f(q) = \cos(\theta_{2,p}) = \frac{\|g\|_q^q}{\sqrt{\|g\|_{2(q-1)}^{2(q-1)}}}
\end{equation}

Taking logarithms:
\begin{equation}
\ln f(q) = q \ln \|g\|_q - (q-1) \ln \|g\|_{2(q-1)}
\end{equation}

For any $l^r$ norm, the derivative with respect to $r$ is:
\begin{align}
\frac{d}{dr} \ln \|g\|_r &= \frac{1}{r}\left(\frac{\sum_i |g_i|^r \ln |g_i|}{\sum_i |g_i|^r} - \ln \|g\|_r\right)
\end{align}

Applying this formula to compute $\frac{d \ln f}{dq}$, after simplification (the $\ln\|g\|$ terms cancel):
\begin{align}
\frac{d \ln f}{dq} &= \frac{\sum_i |g_i|^q \ln |g_i|}{\sum_i |g_i|^q} - \frac{\sum_i |g_i|^{2(q-1)} \ln |g_i|}{\sum_i |g_i|^{2(q-1)}}
\end{align}

\vspace{0.2cm}
Now we show this derivative is non-negative via convexity. Consider the function:
\begin{equation}
\phi(r) = \ln \|g\|_r^r = \ln \sum_i |g_i|^r
\end{equation}

Its first derivative is precisely the weighted average that appears above:
\begin{equation}
\phi'(r) = \frac{\sum_i |g_i|^r \ln |g_i|}{\sum_i |g_i|^r}
\end{equation}

The second derivative, using the quotient rule, is:
\begin{align}
\phi''(r) &= \frac{\sum_i |g_i|^r (\ln |g_i|)^2}{\sum_i |g_i|^r}- \left(\frac{\sum_i |g_i|^r \ln |g_i|}{\sum_i |g_i|^r}\right)^2 \\
&= \text{Var}_{w^{(r)}}[\ln |g|] \geq 0
\end{align}

where $w_i^{(r)} = |g_i|^r/\sum_j |g_j|^r$. Since variance is always non-negative, $\phi$ is convex, hence $\phi'$ is monotonically increasing.

\vspace{0.2cm}
Observe that $\frac{d \ln f}{dq} = \phi'(q) - \phi'(2(q-1))$.

\vspace{0.2cm}
For $q \in (1,2]$, we have $2(q-1) \leq q$ (since $2(q-1) = 2q-2 \leq q$ when $q \leq 2$). \\
Since $\phi'$ is monotonically increasing and $2(q-1) \leq q$:
\vspace{0.2cm}
\begin{align}
\phi'(2(q-1))\leq \phi'(q)\Rightarrow
\frac{d \ln f}{dq} = \phi'(q) - \phi'(2(q-1)) \geq 0
\end{align}

This proves that $\cos(\theta_{2,p})$ is monotonically increasing in $q$ (equivalently, decreasing in $p$).

\vspace{0.4cm}
\textbf{Step 3: Establish the boundary values using limits.}

\vspace{0.2cm}
At $q = 2$ (corresponding to $p = 2$):
\begin{equation}
\cos(\theta_{2,2}) = \frac{\|g\|_2^2}{\|g\|_2^2} = 1
\end{equation}

For the limit as $q \to 1^+$ (corresponding to $p \to \infty$):
\begin{align}
\lim_{q \to 1^+} \cos(\theta_{2,p}) &= \lim_{q \to 1^+} \|g\|_q^q/\sqrt{\|g\|_{2(q-1)}^{2(q-1)}}
\end{align}
\vspace{0.2cm}
As $q \to 1^+$: the numerator approaches $\|g\|_1$, and $2(q-1) \to 0^+$. For the denominator, $\lim_{r \to 0^+} \|g\|_r^r = d$ (the number of non-zero components). Therefore:
\begin{equation}
\cos(\theta_{2,\infty}) = \lim_{q \to 1^+} \cos(\theta_{2,p}) = \frac{\|g\|_1}{\sqrt{d}}
\end{equation}

Since $\cos(\theta_{2,p})$ is monotonically increasing in $q$ (decreasing in $p$), and using:
\begin{equation}
\|g\|_1 = \frac{\|\nabla_x \ell\|_1}{\|\nabla_x \ell\|_2} = \sqrt{\mathtt{PR}_1}
\end{equation}

we conclude:
\begin{align}
\frac{\|g\|_1}{\sqrt{d}} = \sqrt{\frac{\mathtt{PR}_1}{d}}=\cos(\theta_{2,\infty})\leq \cos(\theta_{2,p})
\end{align}
\end{proof}

\newpage
\section{Taylor Expansion of Cosine Similarity}
\label{app:F}

\vspace{0.2cm}
\noindent\textbf{Proposition 3.} \textit{For $q = 1 + \epsilon$ with small $\epsilon$ and normalized gradient components $\pi_i = \frac{|\nabla_x\ell_i|}{\|\nabla_x\ell\|_1}$, the cosine similarity between $l^2$ and $l^p$ perturbations admits the following first-order expansion:}
\begin{equation}
\cos(\theta_{2,p}) = \sqrt{\frac{\mathtt{PR}_1}{d}}\left(1 + \epsilon(H_m - H)\right) + O(\epsilon^2)
\end{equation}

\textit{where $\mathtt{PR}_1 = \left(\frac{\|\nabla_x\ell\|_1}{\|\nabla_x\ell\|_2}\right)^2$ is the participation ratio, $H$ is the Shannon entropy, and $H_m$ is the logarithmic mean entropy.}
\vspace{0.2cm}

\begin{proof}
Starting with the cosine similarity for $q = 1 + \epsilon$:
\begin{equation}
\cos(\theta_{2,p}) = \frac{\|\nabla_x\ell\|_q^q}{\|\nabla_x\ell\|_2\|\nabla_x\ell\|_{2(q-1)}^{q-1}}
\end{equation}

The numerator expands directly as:
\begin{align}
\|\nabla_x\ell\|_q^q &= \sum_i |\nabla_x\ell_i|^{1+\epsilon} \\
&= \|\nabla_x\ell\|_1\Big(1 + \epsilon\sum_i \frac{|\nabla_x\ell_i|}{\|\nabla_x\ell\|_1} \nonumber \\
&\times \log|\nabla_x\ell_i| + O(\epsilon^2)\Big)
\end{align}

For the denominator term $\|\nabla_x\ell\|_{2\epsilon}^\epsilon$:
\begin{align}
\|\nabla_x\ell\|_{2\epsilon}^\epsilon &= \sqrt{d}\,\Big(1 + 2\epsilon\sum_i \frac{\log|\nabla_x\ell_i|}{d} \nonumber + O(\epsilon^2)\Big)^{\frac{1}{2}} \\
&= \sqrt{d}\,\Big(1 + \epsilon\sum_i \frac{\log|\nabla_x\ell_i|}{d} + O(\epsilon^2)\Big)
\end{align}

Combining terms with normalized gradient components $\pi_i$:
\begin{align}
\cos(\theta_{2,p}) &= \frac{\|\nabla_x\ell\|_1}{\|\nabla_x\ell\|_2\sqrt{d}}\Big(1 + \epsilon\big(\sum_i \pi_i\log|\nabla_x\ell_i| \nonumber \\
&- \sum_i \frac{\log|\nabla_x\ell_i|}{d}\big)\Big) + O(\epsilon^2)
\end{align}

The sums relate to entropy measures through:
\begin{align}
\sum_i \pi_i\log|\nabla_x\ell_i| &= -H + \log\|\nabla_x\ell\|_1 \\
\sum_i \frac{\log|\nabla_x\ell_i|}{d} &= -H_m + \log\|\nabla_x\ell\|_1
\end{align}

where:
\begin{align}
H &= -\sum_i \pi_i\log(\pi_i) \\
H_m &= -\log\prod_{i=1}^d(\pi_i)^{\frac{1}{d}}
\end{align}

Therefore:
\begin{equation}
\cos(\theta_{2,p}) = \sqrt{\frac{\mathtt{PR}_1}{d}}\left(1 + \epsilon(H_m - H)\right) + O(\epsilon^2)
\end{equation}

The entropy gap $\Delta H = H_m - H$ is non-negative by Jensen's inequality.
\end{proof}

Treating the norm order as a continuous variable, as in this expansion, also proves useful elsewhere: in neural operator learning, fractional derivative orders learned from data improve accuracy~\cite{mehouachi2026fractional}.

\section{AutoAttack Results}
\label{app:G}

We also report robust accuracy results evaluated with AutoAttack (AA)~\cite{croce2020reliable}. We present the clean (top) and robust (bottom) accuracies (3 seeds) for CIFAR-10 using WRN-28-10, evaluated with AA. AutoAttack gives lower accuracies than PGD-50, for example 42.43\% against 51.31\% for $l^p$-FGSM at $\epsilon=8/255$ (Table~\ref{tab:cifar10-comparison}).

\begin{table}[h!]
\renewcommand{\arraystretch}{1.0}
\caption{CIFAR-10 (WRN-28-10) Clean and AutoAttack Accuracy Evaluation. Results are averaged over three seeds. Clean accuracy (top) and AutoAttack accuracy (bottom).}
\label{tab:cifar10_wrn28_10_auto}
\centering
\resizebox{\columnwidth}{!}{
\begin{tabular}{|c|c|c|c|c|}
\hline
\multicolumn{5}{|c|}{\textbf{CIFAR-10 WRN-28-10 AutoAttack}} \\
\hline
\(\epsilon\cdot255\) & FGSM & RS-FGSM & N-FGSM & $l^p$-FGSM \\
\hline
\multirow{2}{*}{2} & \textbf{90.81\%} {\scriptsize ±0.07} & 90.64\% {\scriptsize ±0.12} & 89.27\% {\scriptsize ±0.21} & 89.02\% {\scriptsize ±0.41} \\
& 74.72\% {\scriptsize ±0.37} & 71.47\% {\scriptsize ±0.44} & 73.14\% {\scriptsize ±0.68} & \textbf{76.14\%} {\scriptsize ±0.62} \\
\hline
\multirow{2}{*}{4} & \textbf{87.86\%} {\scriptsize ±0.23} & 86.58\% {\scriptsize ±0.22} & 86.34\% {\scriptsize ±0.36} & 85.71\% {\scriptsize ±0.53} \\
& 61.58\% {\scriptsize ±0.12} & 54.85\% {\scriptsize ±0.16} & 59.81\% {\scriptsize ±0.27} & \textbf{62.12\%} {\scriptsize ±0.42} \\
\hline
\multirow{2}{*}{8} & \textbf{84.89\%} {\scriptsize ±1.20} & 80.14\% {\scriptsize ±0.88} & 74.73\% {\scriptsize ±0.46} & 79.81\% {\scriptsize ±0.57} \\
& 0.00\% {\scriptsize ±0.00} & 35.77\% {\scriptsize ±0.24} & 41.65\% {\scriptsize ±0.45} & \textbf{42.43\%} {\scriptsize ±0.58} \\
\hline
\multirow{2}{*}{12} & \textbf{80.23\%} {\scriptsize ±0.63} & 61.65\% {\scriptsize ±1.32} & 62.56\% {\scriptsize ±0.73} & 71.12\% {\scriptsize ±0.38} \\
& 0.00\% {\scriptsize ±0.00} & 0.00\% {\scriptsize ±0.00} & 30.17\% {\scriptsize ±1.16} & \textbf{32.13\%} {\scriptsize ±0.71} \\
\hline
\multirow{2}{*}{16} & \textbf{74.61\%} {\scriptsize ±0.19} & 69.20\% {\scriptsize ±0.15} & 52.89\% {\scriptsize ±0.27} & 58.43\% {\scriptsize ±0.48} \\
& 0.00\% {\scriptsize ±0.00} & 0.00\% {\scriptsize ±0.00} & 22.50\% {\scriptsize ±0.89} & \textbf{25.89\%} {\scriptsize ±0.59} \\
\hline
\end{tabular}
}
\end{table}

The comparison encompasses standard FGSM~\cite{goodfellow2014explaining}, RS-FGSM~\cite{wong2020fast}, N-FGSM with (k=2)~\cite{de2022make}, and our proposed adaptive $l^p$-FGSM ($\beta=0.01$).

\begin{figure}[h!]
\centering
\includegraphics[width=\columnwidth]{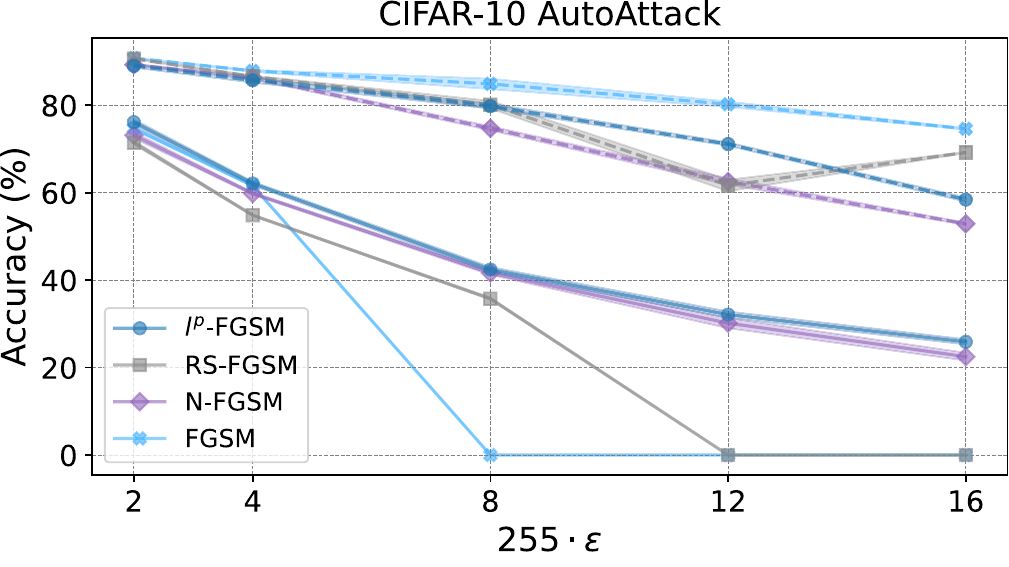}
\caption{Comparative evaluation using AutoAttack on CIFAR-10 with WideResNet-28-10 across different perturbation magnitudes.}
\label{fig:auto}
\end{figure}

The experiments reveal a characteristic pattern of Catastrophic Overfitting (CO) across various perturbation magnitudes ($\epsilon$) for FGSM and RS-FGSM. During CO, models maintain high clean accuracy while their robust accuracy against adversarial attacks deteriorates to near zero. AutoAttack combines several complementary attacks~\cite{croce2020reliable}, which explains its lower accuracies. These results are consistent with our findings regarding the effectiveness of norm selection in preventing CO.

\section{Long-Term Training Evaluation}
\label{app:H}

To rigorously assess the durability and stability of the $l^p$-FGSM method under prolonged training conditions, we conducted an extended training experiment spanning 200 epochs. This experiment utilized the CIFAR-10 and CIFAR-100 datasets with adversarial perturbation norms set at $\epsilon=8/255$ and $\epsilon=16/255$, using ADAM optimizer with a learning rate of 0.001, weight decay of 0.0005, batch size of 128 and WRN-28-10 as an architecture.

\begin{figure*}[!t]
\centering
\includegraphics[width=0.8\textwidth]{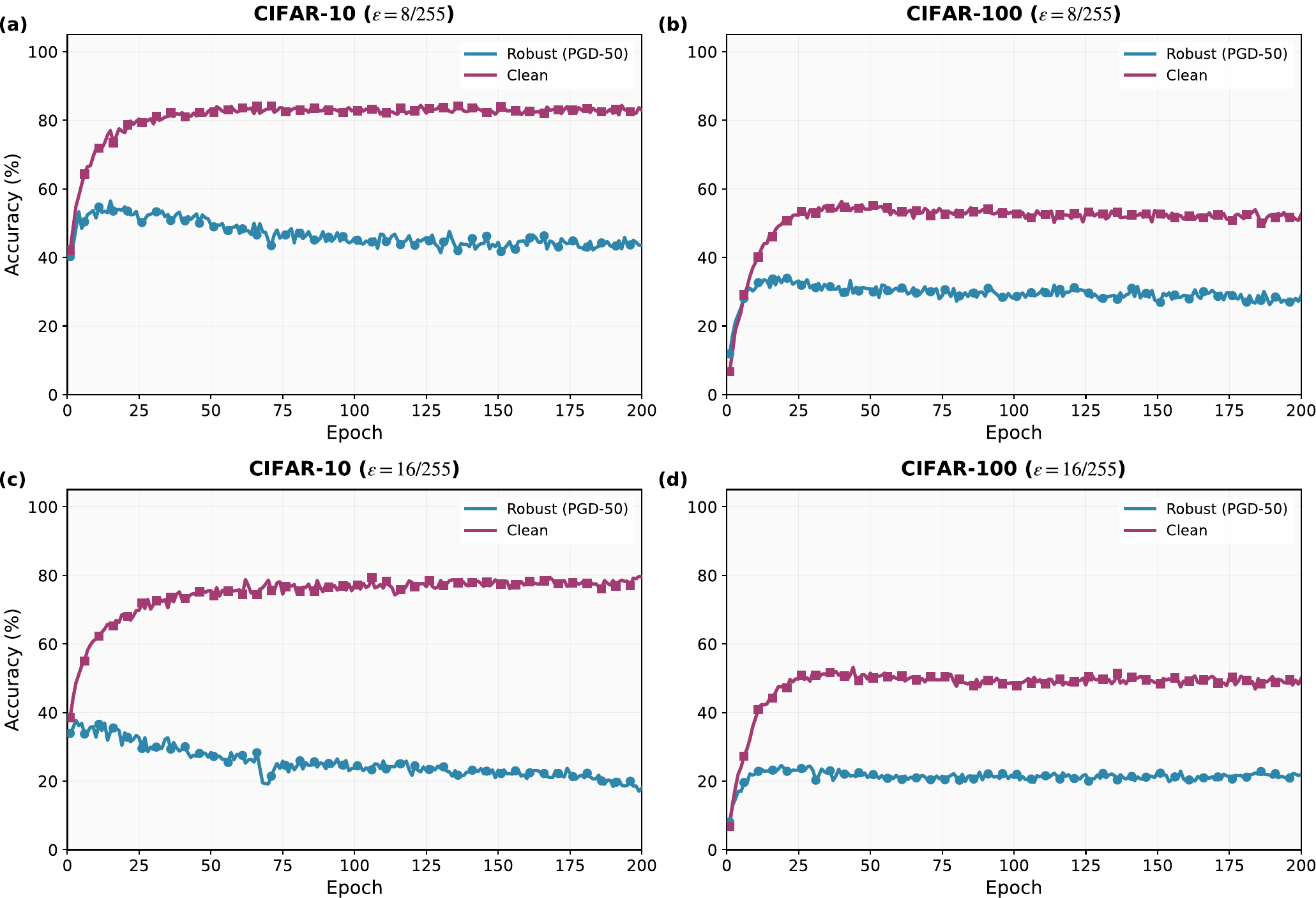}
\caption{Extended training performance of $l^p$-FGSM on CIFAR-10 and CIFAR-100. While Catastrophic Overfitting (CO) was not observed, the experiment highlights the occurrence of robust overfitting over a prolonged training period.}
\label{fig:fig6_25}
\end{figure*}

The results of this long-term training provide insightful observations. Crucially, no instances of Catastrophic Overfitting (CO) were detected throughout the training process, underscoring the robustness of the $l^p$-FGSM approach. However, a slight decrease in robustness, i.e., robust overfitting, occurs. This occurrence warrants early stopping and cyclical learning rates to offset this phenomenon.

\section{$l^p$-FGSM Results Tables}
\label{app:A6}
Tables~\ref{tab:svhn-comparison}, \ref{tab:cifar10-comparison} and~\ref{tab:cifar100-comparison} report, for each $\epsilon$, the clean accuracy (first row) and the PGD-50-10 accuracy (second row) of the methods compared in Section~IV of the main paper, as mean and standard deviation over seeds.

\begin{table*}[!t]
\renewcommand{\arraystretch}{1.0}
\caption{Comparative Analysis of Fast Adversarial Training Methods on SVHN Dataset}
\label{tab:svhn-comparison}
\centering
\resizebox{0.7\textwidth}{!}{
\begin{tabular}{|c|c|c|c|c|c|}
            \hline
            \multicolumn{6}{|c|}{\textbf{SVHN PreAct-18 PGD-50-10}} \\
            \hline
\(\epsilon\cdot255\) & $l^p$-FGSM & RS-FGSM & N-FGSM & GradAlign & ZeroGrad \\
\hline
2 & 94.20\% {\scriptsize ±0.52} & \bv{96.16\%} {\scriptsize ±0.13} & 96.04\% {\scriptsize ±0.24} & 96.01\% {\scriptsize ±0.25} & 96.08\% {\scriptsize ±0.22} \\
& 86.22\% {\scriptsize ±0.22} & 86.17\% {\scriptsize ±0.17} & 86.46\% {\scriptsize ±0.12} & 86.44\% {\scriptsize ±0.15} & \bv{86.47\%} {\scriptsize ±0.17} \\
\hline
4 & 94.16\% {\scriptsize ±0.64} & \bv{95.07\%} {\scriptsize ±0.08} & 94.56\% {\scriptsize ±0.18} & 94.57\% {\scriptsize ±0.24} & 94.83\% {\scriptsize ±0.19} \\
& \bv{77.86\%} {\scriptsize ±0.75} & 71.25\% {\scriptsize ±0.43} & 72.54\% {\scriptsize ±0.21} & 72.18\% {\scriptsize ±0.22} & 71.64\% {\scriptsize ±0.24} \\
\hline
6 & 92.26\% {\scriptsize ±0.65} & \bv{95.16\%} {\scriptsize ±0.48} & 92.27\% {\scriptsize ±0.36} & 92.55\% {\scriptsize ±0.26} & 93.52\% {\scriptsize ±0.24} \\
& \bv{64.12\%} {\scriptsize ±1.27} & 0.00\% {\scriptsize ±0.00} & 58.44\% {\scriptsize ±0.18} & 57.36\% {\scriptsize ±0.27} & 51.77\% {\scriptsize ±0.58} \\
\hline
8 & 91.06\% {\scriptsize ±0.69} & \bv{94.48\%} {\scriptsize ±0.18} & 89.59\% {\scriptsize ±0.48} & 90.16\% {\scriptsize ±0.36} & 92.43\% {\scriptsize ±1.33} \\
& \bv{56.72\%} {\scriptsize ±0.74} & 0.00\% {\scriptsize ±0.00} & 45.64\% {\scriptsize ±0.21} & 43.88\% {\scriptsize ±0.16} & 35.96\% {\scriptsize ±2.78} \\
\hline
10 & 90.76\% {\scriptsize ±1.21} & \bv{93.82\%} {\scriptsize ±0.28} & 86.78\% {\scriptsize ±0.88} & 87.26\% {\scriptsize ±0.73} & 90.36\% {\scriptsize ±0.33} \\
& \bv{45.46\%} {\scriptsize ±1.04} & 0.00\% {\scriptsize ±0.00} & 33.98\% {\scriptsize ±0.48} & 32.88\% {\scriptsize ±0.36} & 21.36\% {\scriptsize ±0.37} \\
\hline
12 & 90.02\% {\scriptsize ±0.38} & \bv{92.72\%} {\scriptsize ±0.56} & 81.49\% {\scriptsize ±1.66} & 84.12\% {\scriptsize ±0.44} & 88.11\% {\scriptsize ±0.47} \\
& \bv{36.88\%} {\scriptsize ±1.09} & 0.00\% {\scriptsize ±0.00} & 26.17\% {\scriptsize ±0.88} & 23.64\% {\scriptsize ±0.42} & 14.16\% {\scriptsize ±0.38} \\
\hline
\end{tabular}
}
\end{table*}

\begin{table*}[!t]
\renewcommand{\arraystretch}{1.0}
\caption{Comparative Analysis of Fast Adversarial Training Methods on CIFAR-10 Dataset}
\label{tab:cifar10-comparison}
\centering
\resizebox{0.7\textwidth}{!}{
\begin{tabular}{|c|c|c|c|c|c|}
            \hline
            \multicolumn{6}{|c|}{\textbf{CIFAR-10 WRN-28-10 PGD-50-10}} \\
            \hline

\(\epsilon\cdot255\) & $l^p$-FGSM & RS-FGSM & N-FGSM & GradAlign & ZeroGrad \\
\hline
2 & 91.12\% {\scriptsize ±0.52} & \bv{92.86\%} {\scriptsize ±0.14} & 92.49\% {\scriptsize ±0.14} & 92.54\% {\scriptsize ±0.13} & 92.62\% {\scriptsize ±0.16} \\
& 80.84\% {\scriptsize ±0.25} & 80.91\% {\scriptsize ±0.14} & \bv{81.42\%} {\scriptsize ±0.34} & 81.32\% {\scriptsize ±0.43} & 81.41\% {\scriptsize ±0.32} \\
\hline
4 & 88.07\% {\scriptsize ±0.34} & \bv{90.74\%} {\scriptsize ±0.23} & 89.64\% {\scriptsize ±0.23} & 89.93\% {\scriptsize ±0.34} & 90.21\% {\scriptsize ±0.22} \\
& 69.62\% {\scriptsize ±0.84} & 68.24\% {\scriptsize ±0.19} & 69.10\% {\scriptsize ±0.27} & \bv{69.80\%} {\scriptsize ±0.48} & 69.21\% {\scriptsize ±0.21} \\
\hline
6 & 83.23\% {\scriptsize ±0.46} & \bv{88.25\%} {\scriptsize ±0.22} & 85.74\% {\scriptsize ±0.32} & 86.94\% {\scriptsize ±0.16} & 86.11\% {\scriptsize ±0.45} \\
& \bv{59.24\%} {\scriptsize ±0.51} & 57.24\% {\scriptsize ±0.19} & 58.26\% {\scriptsize ±0.18} & 59.14\% {\scriptsize ±0.16} & 58.44\% {\scriptsize ±0.19} \\
\hline
8 & 81.67\% {\scriptsize ±0.61} & 83.61\% {\scriptsize ±1.77} & 81.64\% {\scriptsize ±0.35} & 82.16\% {\scriptsize ±0.21} & \bv{84.16\%} {\scriptsize ±0.21} \\
& \bv{51.31\%} {\scriptsize ±0.59} & 0.00\% {\scriptsize ±0.00} & 49.51\% {\scriptsize ±0.27} & 50.12\% {\scriptsize ±0.17} & 48.32\% {\scriptsize ±0.21} \\
\hline
10 & 76.61\% {\scriptsize ±0.58} & \bv{82.17\%} {\scriptsize ±1.48} & 76.94\% {\scriptsize ±0.12} & 79.42\% {\scriptsize ±0.28} & 81.29\% {\scriptsize ±0.73} \\
& \bv{45.87\%} {\scriptsize ±0.68} & 0.00\% {\scriptsize ±0.00} & 42.39\% {\scriptsize ±0.39} & 41.42\% {\scriptsize ±0.52} & 36.18\% {\scriptsize ±0.19} \\
\hline
12 & 72.84\% {\scriptsize ±0.54} & 78.64\% {\scriptsize ±0.74} & 72.18\% {\scriptsize ±0.17} & 73.72\% {\scriptsize ±0.82} & \bv{79.33\%} {\scriptsize ±0.92} \\
& \bv{41.09\%} {\scriptsize ±1.24} & 0.00\% {\scriptsize ±0.00} & 36.82\% {\scriptsize ±0.27} & 35.16\% {\scriptsize ±0.77} & 28.26\% {\scriptsize ±1.81} \\
\hline
14 & 66.58\% {\scriptsize ±0.63} & 73.27\% {\scriptsize ±2.84} & 67.86\% {\scriptsize ±0.46} & 66.41\% {\scriptsize ±0.52} & \bv{78.18\%} {\scriptsize ±0.66} \\
& \bv{38.65\%} {\scriptsize ±0.81} & 0.00\% {\scriptsize ±0.00} & 31.68\% {\scriptsize ±0.68} & 30.85\% {\scriptsize ±0.34} & 18.56\% {\scriptsize ±0.35} \\
\hline
16 & 63.84\% {\scriptsize ±0.76} & 68.68\% {\scriptsize ±2.43} & 56.75\% {\scriptsize ±0.44} & 57.88\% {\scriptsize ±0.74} & \bv{75.43\%} {\scriptsize ±0.89} \\
& \bv{37.16\%} {\scriptsize ±1.22} & 0.00\% {\scriptsize ±0.00} & 25.11\% {\scriptsize ±0.43} & 26.24\% {\scriptsize ±0.43} & 14.66\% {\scriptsize ±0.22} \\
\hline
\end{tabular}
}
\end{table*}

\begin{table*}[!t]
\renewcommand{\arraystretch}{1.0}
\caption{Comparative Analysis of Fast Adversarial Training Methods on CIFAR-100 Dataset}
\label{tab:cifar100-comparison}
\centering
\resizebox{0.7\textwidth}{!}{
\begin{tabular}{|c|c|c|c|c|c|}

            \hline
            \multicolumn{6}{|c|}{\textbf{CIFAR-100 WRN-28-10 PGD-50-10}} \\
            \hline

\(\epsilon\cdot255\) & $l^p$-FGSM & RS-FGSM & N-FGSM & GradAlign & ZeroGrad \\
\hline
2 & 66.42\% {\scriptsize ±0.15} & \bv{72.62\%} {\scriptsize ±0.24} & 71.52\% {\scriptsize ±0.14} & 71.61\% {\scriptsize ±0.23} & 71.64\% {\scriptsize ±0.22} \\
& \bv{55.29\%} {\scriptsize ±0.64} & 51.62\% {\scriptsize ±0.56} & 52.24\% {\scriptsize ±0.35} & 51.51\% {\scriptsize ±0.48} & 52.63\% {\scriptsize ±0.64} \\
\hline
4 & 61.32\% {\scriptsize ±0.34} & \bv{68.27\%} {\scriptsize ±0.21} & 66.51\% {\scriptsize ±0.48} & 67.09\% {\scriptsize ±0.19} & 67.21\% {\scriptsize ±0.18} \\
& \bv{45.73\%} {\scriptsize ±0.46} & 39.56\% {\scriptsize ±0.14} & 39.96\% {\scriptsize ±0.31} & 39.81\% {\scriptsize ±0.48} & 39.61\% {\scriptsize ±0.32} \\
\hline
6 & 58.79\% {\scriptsize ±0.45} & \bv{65.62\%} {\scriptsize ±0.66} & 61.42\% {\scriptsize ±0.63} & 62.86\% {\scriptsize ±0.10} & 63.65\% {\scriptsize ±0.12} \\
& \bv{38.33\%} {\scriptsize ±0.54} & 26.61\% {\scriptsize ±2.79} & 30.99\% {\scriptsize ±0.27} & 32.11\% {\scriptsize ±0.24} & 30.28\% {\scriptsize ±0.51} \\
\hline
8 & 53.46\% {\scriptsize ±0.58} & 54.28\% {\scriptsize ±5.92} & 56.42\% {\scriptsize ±0.65} & 58.55\% {\scriptsize ±0.41} & \bv{60.78\%} {\scriptsize ±0.24} \\
& \bv{32.41\%} {\scriptsize ±1.18} & 0.00\% {\scriptsize ±0.00} & 26.71\% {\scriptsize ±0.68} & 26.97\% {\scriptsize ±0.61} & 23.72\% {\scriptsize ±0.16} \\
\hline
10 & 50.23\% {\scriptsize ±0.42} & 46.18\% {\scriptsize ±4.88} & 51.51\% {\scriptsize ±0.61} & 53.85\% {\scriptsize ±0.73} & \bv{61.11\%} {\scriptsize ±0.39} \\
& \bv{27.12\%} {\scriptsize ±0.76} & 0.00\% {\scriptsize ±0.00} & 23.11\% {\scriptsize ±0.49} & 22.64\% {\scriptsize ±0.61} & 15.15\% {\scriptsize ±0.45} \\
\hline
12 & 47.23\% {\scriptsize ±0.28} & 35.86\% {\scriptsize ±0.27} & 46.42\% {\scriptsize ±0.56} & 46.94\% {\scriptsize ±0.86} & \bv{58.36\%} {\scriptsize ±0.15} \\
& \bv{24.74\%} {\scriptsize ±0.67} & 0.00\% {\scriptsize ±0.00} & 19.32\% {\scriptsize ±0.51} & 19.94\% {\scriptsize ±0.65} & 11.12\% {\scriptsize ±0.66} \\
\hline
14 & 43.18\% {\scriptsize ±0.25} & 24.42\% {\scriptsize ±1.38} & 42.14\% {\scriptsize ±0.36} & 42.63\% {\scriptsize ±0.50} & \bv{56.24\%} {\scriptsize ±0.16} \\
& \bv{22.32\%} {\scriptsize ±1.13} & 0.00\% {\scriptsize ±0.00} & 16.62\% {\scriptsize ±0.44} & 16.96\% {\scriptsize ±0.14} & 8.81\% {\scriptsize ±0.34} \\
\hline
16 & 40.56\% {\scriptsize ±1.64} & 21.47\% {\scriptsize ±5.21} & 38.37\% {\scriptsize ±0.48} & 36.17\% {\scriptsize ±0.45} & \bv{56.42\%} {\scriptsize ±0.29} \\
& \bv{18.41\%} {\scriptsize ±1.42} & 0.00\% {\scriptsize ±0.00} & 14.29\% {\scriptsize ±0.38} & 14.23\% {\scriptsize ±0.26} & 4.92\% {\scriptsize ±0.38} \\
\hline
\end{tabular}
}
\end{table*}

\section{Effects of $\varepsilon$-Softening and Noise Injection}
\label{app:I}

We investigate two key components of our $l^p$-FGSM framework: the $\varepsilon$-softening term from Algorithm 1 (main paper) and the integration of random noise.

The $\varepsilon$-softening term, introduced to maintain Lipschitz continuity in our fixed-point formulation, helps numerical stability by avoiding zero division. Furthermore, there is a contrast with ZeroGrad~\cite{golgooni2021zerograd} that nullifies small gradient components, while our softening ensures gradients maintain minimal non-zero values.

The theoretical motivation behind $\varepsilon$-softening stems from the observation that the fixed-point mapping's contractiveness is particularly sensitive near zero-gradient regions. By introducing a small, non-zero floor to gradient magnitudes, we maintain the desirable theoretical properties of our fixed-point formulation while improving numerical stability~\cite{andriushchenko2020understanding, kim2021understanding}.

For noise integration, following~\cite{wong2020fast}, we can employ a dual-purpose strategy where noise can either serve as input augmentation or initialization for perturbation crafting:
\begin{equation}
\begin{cases}
x_{0}\leftarrow x_{0}+\eta,\ \eta\sim\mathcal{U}[-\epsilon,\epsilon]\\
\delta_{0}\leftarrow\Pi_{\partial B_{p}(\epsilon)}(\eta)
\end{cases}
\end{equation}

These two noise placement approaches can be used independently. The random initialization at boundary $\partial B_p(\epsilon)$ particularly helps when gradient information is near zero. Our implementation differs from previous approaches in two key aspects: first, we project the noise onto the $l^p$ ball boundary rather than using uniform sampling, and second, we reuse the same noise vector for both input augmentation and initialization, reducing computational overhead~\cite{shafahi2019adversarial}.

\begin{figure}[h!]
\centering
\includegraphics[width=\columnwidth]{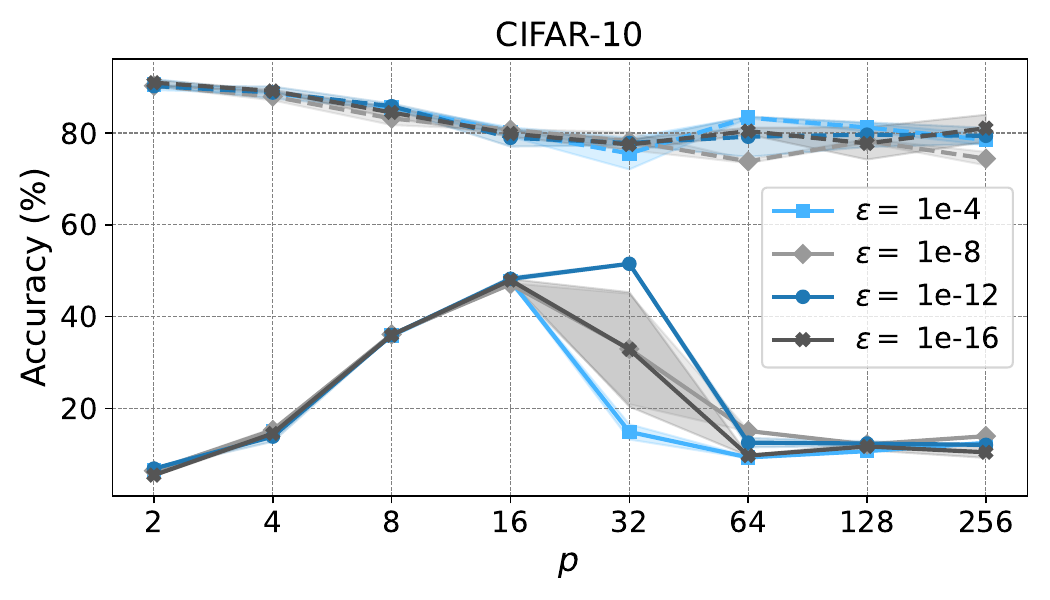}
\vskip 0.1in
\includegraphics[width=\columnwidth]{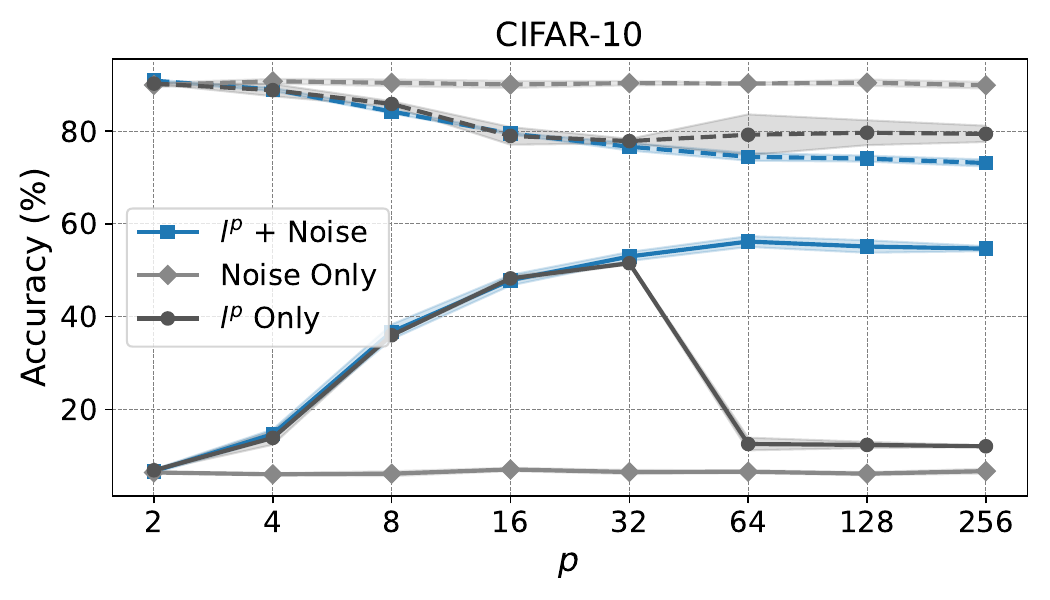}
\caption{Analysis of $\varepsilon$-softening and noise effects on CIFAR-10 using WideResNet-28-10 against PGD-50 ($\epsilon=8/255$). Top: Effect of $\varepsilon$-softening on clean (dashed) and adversarial (solid) accuracy for various $p$ values. Optimal $\varepsilon$ enhances stability against CO. Bottom: Synergistic effects of noise injection showing improved robustness against CO and enhanced overall accuracy. In these runs, both components help prevent catastrophic overfitting while maintaining competitive performance.}
\label{fig:appendix}
\end{figure}

Even though the main paper does not use any noise, the synergistic relationship between $\varepsilon$-softening and noise injection becomes apparent in their complementary effects on training stability. While $\varepsilon$-softening provides consistent gradient behavior, noise injection helps explore the loss landscape more effectively. This combination proves particularly effective in preventing the gradient collapse often associated with CO~\cite{andriushchenko2020understanding}.

Our extensive experiments on CIFAR-10 with WideResNet-28-10 (Figure~\ref{fig:appendix}) demonstrate that both components contribute meaningfully to the algorithm's performance. The $\varepsilon$-softening exhibits an optimal range where it enhances stability without compromising accuracy, while noise injection provides complementary benefits in preventing CO and improving overall robustness.

The combination of these techniques may also allow more aggressive training schedules~\cite{wong2020fast, rice2020overfitting}. These findings suggest promising directions for future research in stabilizing adversarial training in conjunction with our adaptive $l^p$-FGSM.

\section{Entropy Gap and $\mathtt{PR}_1$ for $l^\infty$ vs $l^p$}
\label{app:K}
\begin{figure}[t!]
\centering
\includegraphics[width=\columnwidth]{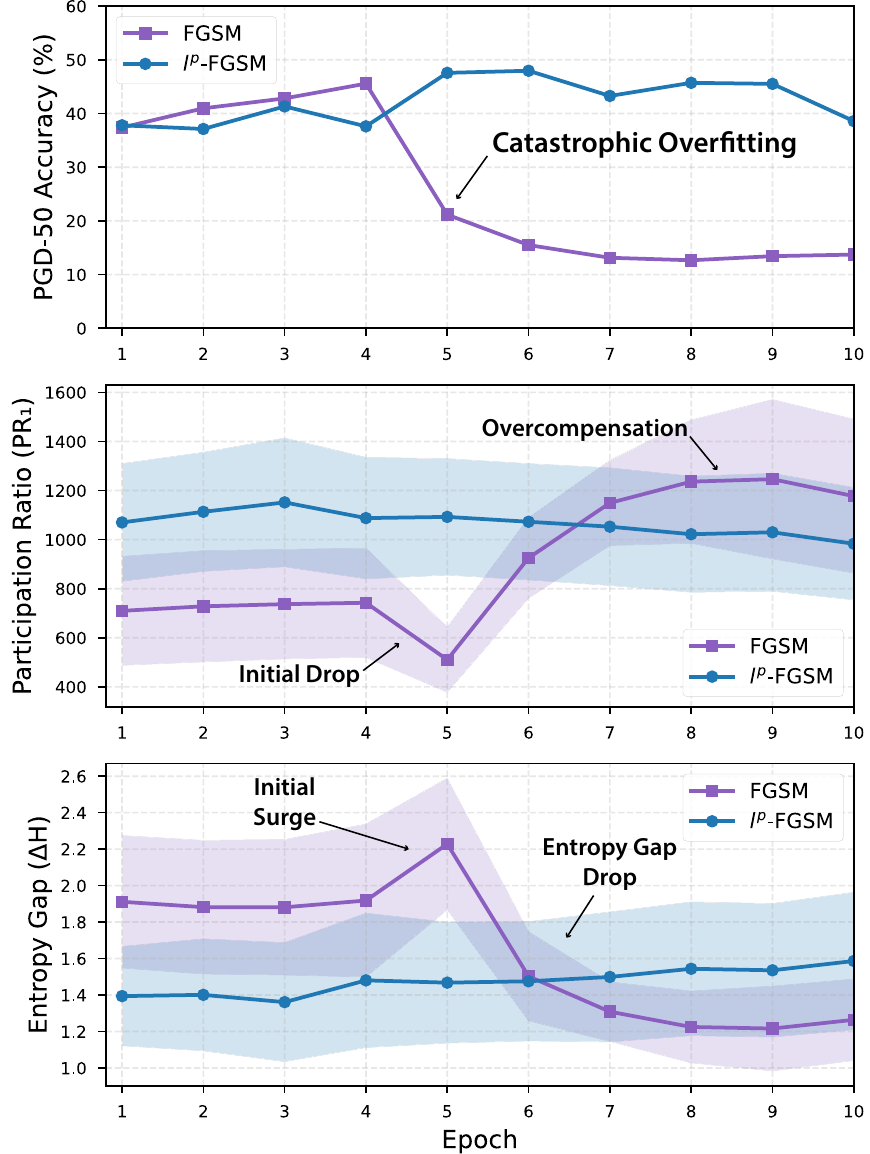}
\caption{Evolution of Participation Ratios ($\mathtt{PR_1}$) and entropy gap during training with and without $l^p$-FGSM. Sharp patterns in these metrics align with the onset of Catastrophic Overfitting (CO), highlighting the link between gradient concentration and adversarial vulnerability. Same experimental setting as in the main paper (CIFAR-10, WRN-28-10, ADAM, learning rate 0.001, batch size 128).}
\label{fig:entropy_gap}
\end{figure}

Our preliminary analysis suggests that gradient concentration metrics (Participation Ratio and entropy gap) exhibit notable changes that appear to coincide with the onset of Catastrophic Overfitting. As shown in Figure~\ref{fig:entropy_gap}, these metrics display an interesting pattern that warrants further investigation: a moderate increase, followed by a drop, and then what appears to be a compensatory response. While more extensive experimentation is needed to fully validate these observations, the pattern is consistent across multiple experimental runs.

The adaptation of Participation Ratio ($\mathtt{PR}$) from quantum mechanics~\cite{anderson1958absence, feynman2011feynman} to the adversarial training context as $\mathtt{PR}_1$ represents a novel approach to quantifying gradient behavior. In quantum systems, PR measures the effective number of states occupied by an electron; similarly, our $\mathtt{PR}_1$ aims to capture the effective dimensionality of gradient information. The entropy gap metric offers a complementary perspective, potentially providing insights into how information is distributed across gradient dimensions.

The observed pattern, an initial increase, decline, and subsequent adjustment, may offer preliminary insights into the dynamics preceding CO. This behavior could potentially reflect the model's changing gradient geometry as it negotiates the complex loss landscape during adversarial training. The initial increase in both $\mathtt{PR}_1$ and entropy gap might suggest a temporary distribution of gradient information before concentration occurs.

By leveraging these metrics during training, our adaptive norm selection approach aims to detect potential instabilities and adjust accordingly. While our current results are promising, we acknowledge that the full relationship between these information-theoretic measures and adversarial robustness requires deeper exploration.

These initial findings provide support for our theoretical framework connecting gradient geometry to norm selection, suggesting that the $l^p$-FGSM approach may effectively mitigate CO without requiring additional techniques like gradient alignment or noise injection. Future work could explore these connections more thoroughly, potentially yielding broader insights into neural network behavior under adversarial constraints.

\section{Comparison with Recent Fast Adversarial Training Methods}
\label{app:fastat}

\noindent\textbf{Protocol.} We evaluate adaptive $l^p$-FGSM inside the FastAT Benchmark~\cite{pan2026fastat}, which implements fast adversarial training methods in one codebase and evaluates them under one protocol. We train $l^p$-FGSM with the benchmark's released code and change only the attack step of its FGSM-AT implementation. The settings are those of the benchmark: ResNet-18; 49{,}000 training and 1{,}000 validation images; 100 epochs of SGD with a one-cycle learning rate (peak 0.1), momentum 0.9, weight decay $5\times10^{-4}$ and batch size 128; label smoothing 0.4; and weight averaging with decay 0.9995. The checkpoint with the highest PGD-10 accuracy on the validation images, with or without weight averaging, is tested on the 10{,}000 test images with PGD-10, PGD-20 and PGD-50 (step $2/255$), AutoAttack (standard version)~\cite{croce2020reliable} and the CR attack, all at $\epsilon=8/255$. We use $\beta=0.04$, selected among $\{0.01, 0.02, 0.03, 0.04, 0.06\}$ on the CIFAR-10 validation images, and the same value for CIFAR-100. These experiments use a different architecture and schedule from those of Section~IV of the main paper (WRN-28-10, 30 epochs).

\medskip
\noindent\textbf{Results.} Tables~\ref{tab:fastat10} and~\ref{tab:fastat100} report the results next to all methods of the benchmark. On CIFAR-10, $l^p$-FGSM alone reaches $46.77\pm0.14\%$ AutoAttack accuracy, against 37.28\% for FGSM-AT, whose step it replaces, without catastrophic overfitting on any seed and at 1.02 times its cost. Its combination with a noisy start and a consistency term, below, trains stably on every seed of both datasets.

\begin{figure*}[!t]
\centering
\includegraphics[width=0.6\linewidth]{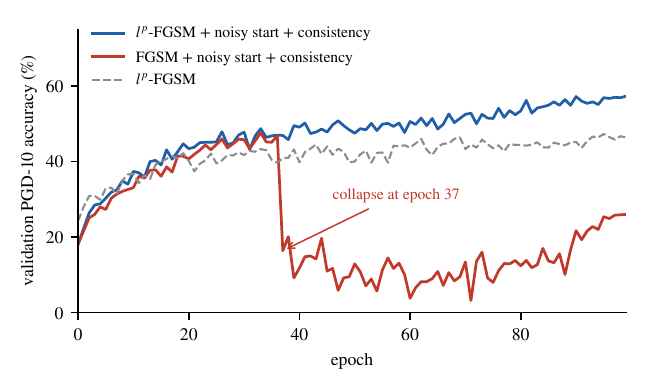}
\caption{Validation PGD-10 accuracy during training on CIFAR-10, seed 0, under the FastAT Benchmark protocol. With the same noisy start and consistency term, the FGSM step collapses at epoch 37 on this seed, while the adaptive $l^p$ step does not. The dashed line is $l^p$-FGSM without these two ingredients.}
\label{fig:control}
\end{figure*}

\medskip
\noindent\textbf{Combination with a noisy start and a consistency term.} The strongest single-step methods of the benchmark share two ingredients: a starting point other than the clean image, and a consistency term between the logits at the starting point and at the adversarial point. FGSM-PGI~\cite{jia2022prior} and FGSM-UAP~\cite{pan2024uap} use the mean squared difference of the logits with weight 10, and LIET~\cite{pan2025liet} a Jensen--Shannon version. We add both ingredients to $l^p$-FGSM. The starting point is $x_0=\operatorname{clip}_{[0,1]}(x+\eta)$ with $\eta\sim\mathcal{U}[-\epsilon,\epsilon]^d$; the $l^p$ step is taken from $x_0$, and the result is clipped to $[0,1]$ but not projected back onto the $\epsilon$-ball, as in N-FGSM~\cite{de2022make}. The training loss is
\[
\ell\big(f(x_{\rm adv}),y\big)+10\cdot\frac{1}{C}\big\|f(x_{\rm adv})-f(x_0)\big\|_2^2,
\]
averaged over the batch, where $f$ gives the $C$ logits and both terms are differentiated. We keep $\beta=0.04$, as no other value improved PGD-50 accuracy on the validation images by one point or more. On CIFAR-10 this reaches $49.62\pm0.18\%$ AutoAttack accuracy, within 0.5 points of the best single-step methods of the benchmark (NuAT, 50.10\%; LIET, 49.99\%), and the highest PGD-50 accuracy of all its methods (56.12\%), at 1.42 times the cost of FGSM-AT. On CIFAR-100 it reaches $26.40\pm0.11\%$, second among the single-step methods after LIET (26.64\%), at 1.45 times. As a control, we train the same recipe with the FGSM step. It suffers catastrophic overfitting on one of the three seeds (Figure~\ref{fig:control}): on seed 0, its validation PGD-10 accuracy falls from 47\% to 16\% at epoch 37, and its best checkpoint, from epoch 33, reaches 39.21\% AutoAttack accuracy. On the other two seeds it reaches 49.34\% AutoAttack accuracy, with a clean accuracy of 75.39\%, about 3 points below the $l^p$ step.

\begin{table*}[!t]
\centering
\small
\caption{CIFAR-10 under the FastAT Benchmark protocol (ResNet-18, $\epsilon=8/255$). Accuracy in \%, mean $\pm$ standard deviation over seeds 0, 1 and 2. Rows above the line are the values published in~\cite{pan2026fastat}; rows below are ours, trained with the benchmark code and protocol. Cost: training time relative to FGSM-AT (published times for the benchmark rows; same code and GPU for ours). Methods sorted by AutoAttack accuracy, also shown in Figure~\ref{fig:aa}.}
\label{tab:fastat10}
\begin{tabular}{lcccccc}
\hline
Method & Clean & PGD-10 & PGD-50 & AutoAttack & CR & Cost \\
\hline
PGD-AT-WA~\cite{madry2017towards,izmailov2018averaging} & $82.36\pm0.32$ & $55.05\pm0.34$ & $54.17\pm0.17$ & $50.49\pm0.15$ & $50.51\pm0.16$ & 3.52 \\
NuAT~\cite{sriramanan2021towards} & $80.98\pm0.91$ & $55.33\pm0.14$ & $54.63\pm0.25$ & $50.10\pm0.18$ & $50.11\pm0.19$ & 1.47 \\
LIET~\cite{pan2025liet} & $80.68\pm0.51$ & $56.69\pm0.18$ & $55.97\pm0.19$ & $49.99\pm0.15$ & $49.93\pm0.15$ & 1.31 \\
FGSM-PGI~\cite{jia2022prior} & $79.72\pm0.63$ & $56.43\pm0.18$ & $55.82\pm0.21$ & $49.89\pm0.08$ & $49.89\pm0.12$ & 1.33 \\
FGSM-FUAP~\cite{pan2024uap} & $80.64\pm1.28$ & $56.35\pm0.31$ & $55.75\pm0.38$ & $49.73\pm0.12$ & $49.71\pm0.12$ & 1.47 \\
FGSM-CUAP~\cite{pan2024uap} & $80.76\pm2.10$ & $56.30\pm0.02$ & $55.58\pm0.22$ & $49.55\pm0.22$ & $49.54\pm0.22$ & 1.36 \\
FGSM-UAP~\cite{pan2024uap} & $80.61\pm0.48$ & $56.49\pm0.17$ & $55.77\pm0.13$ & $49.44\pm0.14$ & $49.37\pm0.15$ & 1.28 \\
GAT~\cite{sriramanan2020guided} & $85.10\pm0.13$ & $54.91\pm0.06$ & $53.81\pm0.14$ & $49.15\pm0.15$ & $49.15\pm0.17$ & 1.44 \\
FGSM-RS-CS~\cite{zhao2023fast} & $80.54\pm2.06$ & $53.99\pm0.23$ & $53.12\pm0.29$ & $48.64\pm0.38$ & $48.70\pm0.34$ & 0.91 \\
FGSM-PCO~\cite{wang2024preventing} & $84.69\pm0.32$ & $55.04\pm0.25$ & $54.18\pm0.21$ & $48.52\pm0.32$ & $48.48\pm0.34$ & 1.50 \\
GradAlign~\cite{andriushchenko2020understanding} & $79.39\pm1.97$ & $53.57\pm0.44$ & $52.87\pm0.29$ & $47.99\pm0.62$ & $48.02\pm0.61$ & 2.81 \\
N-FGSM~\cite{de2022make} & $78.59\pm2.34$ & $53.64\pm0.25$ & $53.07\pm0.28$ & $47.77\pm0.41$ & $47.84\pm0.41$ & 0.97 \\
N-AAER~\cite{lin2023eliminating} & $77.67\pm0.48$ & $52.84\pm0.16$ & $52.37\pm0.15$ & $47.38\pm0.18$ & $47.48\pm0.20$ & 0.93 \\
ZeroGrad~\cite{golgooni2021zerograd} & $75.85\pm3.07$ & $51.59\pm2.06$ & $51.06\pm1.98$ & $45.94\pm2.19$ & $46.04\pm2.20$ & 0.96 \\
FGSM-MEP-CS~\cite{zhao2023fast} & $83.55\pm3.14$ & $51.11\pm0.54$ & $49.80\pm0.36$ & $45.35\pm0.81$ & $45.37\pm0.78$ & 1.07 \\
Free-AT~\cite{shafahi2019adversarial} & $75.20\pm0.43$ & $49.32\pm0.22$ & $48.83\pm0.26$ & $44.34\pm0.17$ & $44.32\pm0.24$ & 0.99 \\
PGD-AT~\cite{madry2017towards} & $85.35\pm0.14$ & $48.35\pm0.18$ & $46.31\pm0.10$ & $44.33\pm0.08$ & $44.50\pm0.13$ & 3.48 \\
FGSM-RS~\cite{wong2020fast} & $77.66\pm4.65$ & $49.15\pm5.09$ & $48.27\pm5.33$ & $43.09\pm5.33$ & $43.14\pm5.29$ & 0.95 \\
SSAT~\cite{kim2021understanding} & $84.80\pm2.24$ & $52.40\pm0.45$ & $51.46\pm0.57$ & $38.33\pm1.55$ & $38.32\pm1.53$ & 1.06 \\
FGSM-AT~\cite{goodfellow2014explaining} & $69.17\pm0.95$ & $42.99\pm0.33$ & $42.49\pm0.27$ & $37.28\pm0.30$ & $37.27\pm0.27$ & 1.00 \\
ELLE~\cite{abadrocamora2024efficient} & $77.78\pm8.82$ & $50.74\pm3.46$ & $47.19\pm8.82$ & $32.46\pm25.27$ & $39.77\pm12.77$ & 1.47 \\
\hline
$l^p$-FGSM & $82.18\pm0.85$ & $52.84\pm0.17$ & $51.97\pm0.25$ & $46.77\pm0.14$ & $46.77\pm0.14$ & 1.02 \\
$l^p$-FGSM + noisy start + consistency & $78.23\pm0.27$ & $56.62\pm0.06$ & $56.12\pm0.05$ & $49.62\pm0.18$ & $49.66\pm0.18$ & 1.42 \\
FGSM + noisy start + consistency (control) & $71.14\pm7.40$ & $53.00\pm5.98$ & $52.54\pm5.72$ & $45.96\pm5.85$ & $45.96\pm5.87$ & 1.40 \\
\hline
\end{tabular}
\end{table*}

\begin{table*}[!t]
\centering
\small
\caption{CIFAR-100 under the FastAT Benchmark protocol (ResNet-18, $\epsilon=8/255$). Accuracy in \%, mean $\pm$ standard deviation over seeds 0, 1 and 2. Rows above the line are the values published in~\cite{pan2026fastat}; rows below are ours, trained with the benchmark code and protocol. Cost: training time relative to FGSM-AT (published times for the benchmark rows; same code and GPU for ours). Methods sorted by AutoAttack accuracy, also shown in Figure~\ref{fig:aa}. On CIFAR-100, $l^p$-FGSM alone with $\beta=0.04$ overfits catastrophically late in training, after the checkpoint selected on the validation images; a larger $\beta$ should be used on this dataset.}
\label{tab:fastat100}
\begin{tabular}{lcccccc}
\hline
Method & Clean & PGD-10 & PGD-50 & AutoAttack & CR & Cost \\
\hline
PGD-AT-WA~\cite{madry2017towards,izmailov2018averaging} & $57.62\pm0.34$ & $32.31\pm0.16$ & $31.90\pm0.14$ & $26.90\pm0.17$ & $26.87\pm0.17$ & 3.39 \\
LIET~\cite{pan2025liet} & $50.47\pm0.69$ & $32.64\pm0.53$ & $32.44\pm0.51$ & $26.64\pm0.13$ & $26.63\pm0.14$ & 1.13 \\
FGSM-PGI~\cite{jia2022prior} & $56.61\pm0.34$ & $32.18\pm0.39$ & $31.85\pm0.23$ & $26.29\pm0.16$ & $26.31\pm0.12$ & 1.28 \\
FGSM-RS-CS~\cite{zhao2023fast} & $54.79\pm1.07$ & $32.42\pm0.29$ & $32.08\pm0.29$ & $26.20\pm0.26$ & $26.20\pm0.26$ & 0.91 \\
FGSM-FUAP~\cite{pan2024uap} & $56.29\pm1.13$ & $31.76\pm0.31$ & $31.36\pm0.46$ & $25.95\pm0.08$ & $25.94\pm0.05$ & 1.45 \\
FGSM-CUAP~\cite{pan2024uap} & $55.12\pm1.80$ & $31.65\pm0.26$ & $31.40\pm0.31$ & $25.85\pm0.50$ & $25.83\pm0.49$ & 1.38 \\
FGSM-UAP~\cite{pan2024uap} & $57.84\pm0.28$ & $31.43\pm0.24$ & $31.07\pm0.31$ & $25.84\pm0.24$ & $25.82\pm0.23$ & 1.28 \\
GradAlign~\cite{andriushchenko2020understanding} & $54.61\pm0.75$ & $31.21\pm0.05$ & $30.92\pm0.02$ & $25.37\pm0.14$ & $25.39\pm0.15$ & 2.71 \\
ZeroGrad~\cite{golgooni2021zerograd} & $57.25\pm0.98$ & $30.70\pm0.26$ & $30.15\pm0.32$ & $25.00\pm0.14$ & $24.95\pm0.14$ & 0.98 \\
N-FGSM~\cite{de2022make} & $53.22\pm2.73$ & $30.58\pm0.25$ & $30.35\pm0.25$ & $24.86\pm0.46$ & $24.89\pm0.49$ & 1.00 \\
FGSM-MEP-CS~\cite{zhao2023fast} & $59.60\pm1.88$ & $31.42\pm0.12$ & $30.86\pm0.11$ & $24.86\pm0.28$ & $24.80\pm0.26$ & 1.02 \\
N-AAER~\cite{lin2023eliminating} & $51.95\pm0.75$ & $27.37\pm1.38$ & $26.95\pm1.43$ & $22.74\pm1.17$ & $22.81\pm1.18$ & 0.84 \\
GAT~\cite{sriramanan2020guided} & $65.52\pm0.34$ & $27.66\pm0.16$ & $26.67\pm0.16$ & $22.06\pm0.08$ & $21.97\pm0.07$ & 1.42 \\
Free-AT~\cite{shafahi2019adversarial} & $48.98\pm0.19$ & $27.39\pm0.31$ & $27.19\pm0.34$ & $21.76\pm0.25$ & $21.74\pm0.27$ & 0.96 \\
PGD-AT~\cite{madry2017towards} & $59.11\pm0.39$ & $26.68\pm0.09$ & $25.82\pm0.09$ & $21.44\pm0.23$ & $20.81\pm0.30$ & 3.24 \\
FGSM-RS~\cite{wong2020fast} & $49.00\pm3.64$ & $27.50\pm2.54$ & $27.32\pm2.51$ & $21.42\pm2.24$ & $21.46\pm2.24$ & 1.00 \\
ELLE~\cite{abadrocamora2024efficient} & $47.48\pm3.45$ & $26.68\pm2.45$ & $26.34\pm2.53$ & $21.21\pm2.37$ & $21.23\pm2.37$ & 1.42 \\
FGSM-PCO~\cite{wang2024preventing} & $64.45\pm1.08$ & $26.32\pm0.42$ & $25.69\pm0.28$ & $19.97\pm0.43$ & $19.91\pm0.39$ & 1.36 \\
SSAT~\cite{kim2021understanding} & $66.79\pm1.11$ & $24.88\pm0.52$ & $23.10\pm0.56$ & $18.83\pm0.73$ & $18.79\pm0.79$ & 1.03 \\
FGSM-AT~\cite{goodfellow2014explaining} & $39.50\pm0.77$ & $21.42\pm0.55$ & $21.18\pm0.54$ & $16.62\pm0.54$ & $16.58\pm0.50$ & 1.00 \\
NuAT~\cite{sriramanan2021towards} & $52.63\pm14.39$ & $25.37\pm1.81$ & $21.63\pm1.70$ & $14.41\pm3.29$ & $17.66\pm0.42$ & 1.49 \\
\hline
$l^p$-FGSM & $57.06\pm1.75$ & $30.30\pm0.33$ & $29.86\pm0.24$ & $24.27\pm0.57$ & $24.22\pm0.57$ & 1.04 \\
$l^p$-FGSM + noisy start + consistency & $53.11\pm1.31$ & $31.96\pm0.21$ & $31.71\pm0.20$ & $26.40\pm0.11$ & $26.41\pm0.13$ & 1.45 \\
\hline
\end{tabular}
\end{table*}

\medskip
\noindent\textbf{Cost.} The benchmark does not state its GPU model, so we report training time relative to FGSM-AT. For our rows, FGSM-AT was trained with the same code on the same GPU (NVIDIA A100, 10.0 s per epoch on CIFAR-10).

\begin{figure*}[!t]
\centering
\includegraphics[width=\linewidth]{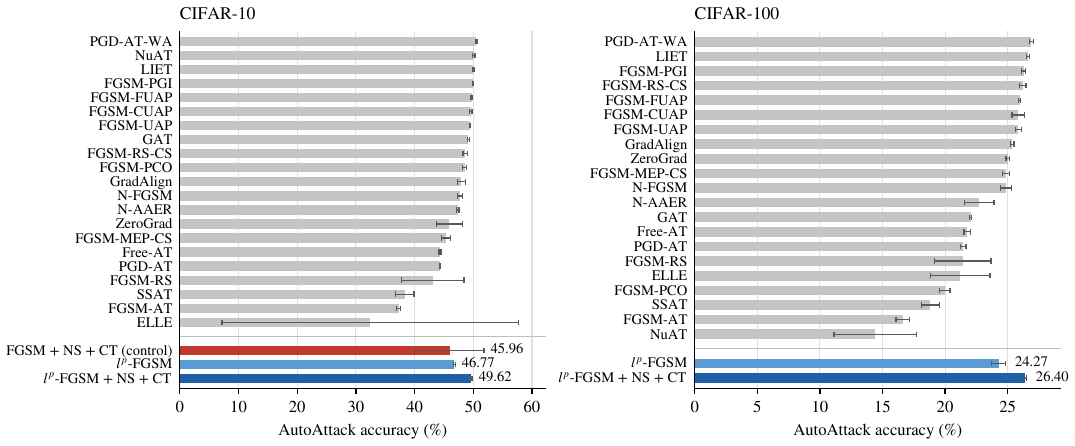}
\caption{AutoAttack accuracy of every method of Tables~\ref{tab:fastat10} (CIFAR-10, left) and~\ref{tab:fastat100} (CIFAR-100, right), mean and standard deviation over three seeds. Gray: values published in~\cite{pan2026fastat}; at the bottom, in color: ours. NS: noisy start; CT: consistency term.}
\label{fig:aa}
\end{figure*}

\end{document}